\documentclass[postprint,12pt]{elsarticle}%

\usepackage{natbib}
\usepackage[utf8]{inputenc}
\usepackage{graphicx}
\usepackage{amsmath}
\usepackage{amsfonts}
\usepackage{amssymb}
\usepackage{multirow}
\usepackage{array}
\usepackage{booktabs}
\usepackage{cancel}
\usepackage[pdftex]{hyperref}
\usepackage{amsthm}
\usepackage{multicol}
\usepackage{enumerate} 
\usepackage{color}
\hypersetup{colorlinks=true,allcolors=black} 
\usepackage[dvipsnames,svgnames,x11names,hyperref,table]{xcolor}
\usepackage{epsfig}
\usepackage{array}
\usepackage{colortbl}
\usepackage{float}
\usepackage{lineno}
\usepackage{amsmath}
\usepackage{subfigure} 
\usepackage{comment}

\newcommand{\base}{BL}
\newcommand{\basel}{BL$_L$}
\newcommand{\basep}{BL$_Q$}
\newcommand{\dsil}{DSIL}
\newcommand{\dsilm}{DSIL$_{\phi}$}
\newcommand{\dsilk}{DSIL$_{K_{\phi}}$}
\newcommand{\dsilkq}{DSIL$_{K_Q}$}
\newcommand{\mplc}{MPLC}
\newcommand{\sr}{SR}
\newcommand{\sre}{SR$_E$}
\newcommand{\srm}{SR$_M$}
\makeatletter
\def\ps@pprintTitle{%
 \let\@oddhead\@empty
 \let\@evenhead\@empty
 \def\@oddfoot{\footnotesize\itshape
       Postprint submitted to \ifx\@journal\@empty Elsevier
       \else\@journal\fi\hfill}
 \let\@evenfoot\@oddfoot}
\makeatother
\journal{Neurocomputing}

\begin{document}
\begin{frontmatter}


\title{Direct side information learning for zero-shot regression}

\author[arcelor]{Miriam Fdez-Díaz \corref{cor1}}
\ead{miriam.fernandezdiaz@arcelormittal.com}
\author[uniovi]{Elena Monta\~n\'es}
\ead{montaneselena@uniovi.es}
\author[uniovi]{Jos\'e Ram\'on Quevedo}
\ead{quevedo@uniovi.es}

\cortext[cor1]{Corresponding author}

\address[arcelor]{ArcelorMittal (Spain)}
\address[uniovi]{Artificial Intelligence Center. University of Oviedo at Gij\'on, 33204 Asturias,  Spain \texttt{http://www.aic.uniovi.es}}

\begin{abstract}
Zero-shot learning provides models for targets for which instances are not available, commonly called unobserved targets. The availability of target side information becomes crucial in this context in order to properly induce models for these targets. The literature is plenty of strategies to cope with this scenario, but specifically designed on the basis of a zero-shot classification scenario, mostly in computer vision and image classification, but they are either not applicable or easily extensible for a zero-shot regression framework for which a continuos value is required to be predicted rather than a label. In fact, there is a considerable lack of methods for zero-shot regression in the literature. Two approaches for zero-shot regression that work in a two-phase procedure were recently proposed. They first learn the observed target models through a classical regression learning ignoring the target side information. Then, they aggregate those observed target models afterwards exploiting the target side information and the models for the unobserved targets are induced. Despite both have shown quite good performance because of the different treatment they grant to the common features and to the side information, they exploit features and side information separately, avoiding a global optimization for providing the unobserved target models. The proposal of this paper is a novel method that jointly takes features and side information in a one-phase learning process, but treating side information properly and in a more deserving way than as common features. A specific kernel that properly merges features and side information is proposed for this purpose resulting in a novel approach that exhibits better performance over both artificial and real datasets.

\end{abstract}
\begin{keyword}
Side information \sep zero-shot regression \sep kernel
\end{keyword}
\end{frontmatter}

\section{Introduction} \label{Introduction}
Improving predictions of air pollutants in meteorological stations makes arise the research of this work, in particular some damaging pollutants (NO$_2$, PST, NO, SO$_2$, CO, O$_3$) collected in the Principality of Asturias, Spain. There are several factors that hardly condition the concentration of these pollutants. Both weather conditions (temperature, humidity, pressure...) and the activities around the meteorological stations (industry, leisure centers, residential areas, power plants, administrative buildings....) seem to be the most influential. However, there are from different nature. On the one hand, weather conditions depend on the climate and vary along the day, weeks, months and seasons.  On the other hand, activities around the stations are constant for each station, hardly vary along the time and are known beforehand even if weather conditions have not been collected yet. Some studies ignore the information about activities in the surroundings \cite{kumar2018evolving}. Just some studies consider both kinds of factors, but they treat them separately and in a different manner. An advanced work considers activities in the surroundings to perform a previous split of the stations resulting in a set of models, one per station or group of stations, where just weather conditions are taken as features \cite{masmoudi2020machine}. Also, taking pollutant values from the nearest neighbor station \cite{delavar2019novel} is a preliminary work of including information about the stations. A prior classification of the stations taking into account roads, traffic flow and the area (urban, suburban or industrial) where the station is located \cite{murillo2019forecasting} is another advanced strategy that considers the surroundings of the stations. However, this information, called side (or privileged) information \cite{guo2017synthesizing,kodirov2015unsupervised,lampert2013attribute,romera2015embarrassingly, waegeman2019multi}, can go away along if it is exploited properly, for instance, including it in the learning process, in order to improve the pollutant predictions. Side information is neither features nor targets; it actually constitutes additional and, prior information about how targets (a pollutant measure in stations) are related to features (weather conditions). Also, treating surrounding information about stations as side information allows us to make predictions over potential future locations of meteorological stations, for which only side information is available. This perspective enables us to state the problem as a zero-shot regression learning task. 

Zero-shot \cite{wang2019asurvey} is a kind of learning that tries to provide predictions for targets devoid of instances. These targets are commonly called unobserved targets. The lack of instances for the unobservable targets really complicates the task of providing promising models to these targets, since only instances for observable targets are available. Then, one must draw on alternative resources to fill the gap. This is the point where side information is crucial and comes into play. Side information is commonly found in form of features \cite{kang2017incorporating, palatucci2009zero}, as it will be in our case.  But depending on the context, it can be found in different formats, for instance in a hierarchy \cite{stefan2019combining}, in a response prediction \cite{menon2011response} or in a structure \cite{jacob2008protein}. Besides, sometimes it is multimodal and heterogeneous, as it happens in recommender systems \cite{liu2019recommender}. In any case, all their special properties make side information worth exploiting in a way far from the way the features are exploited \cite{fdez2022target}. Its prominence is such in the zero-shot framework that great efforts have been made in the literature to collect it \cite{palatucci2009zero, qiao2016less}, or even, to extract or to learn it \cite{farias2019learning, qiam2017alternative}. 

Lately, zero-shot learning has been a booming topic due to the increasing number of applications that demand predicting unobservable targets. This can be, for instance, the case of COVID-19 diagnosis classification, a recent topic that impacts the world notoriously \cite{rezaei2020zero}. Natural language processing \cite{gu2019improved,johnson2017google,levy2017zero,ma2016label, shigeto2015ridge}, videos \cite{elhoseiny2016zero,gan2017deck,gao2019know, mettes2017spatial,xu2016video,xu2017transductive}, mobile and wireless security \cite{robyns2017physical},  emoji predictions \cite{cappallo2015image2emoji,zhan2019zero}, human activity recognition \cite{wang2017zero}, or neuroimaging data \cite{caceres2017feature} are other applications. However, the great majority of applications of zero-shot learning concerns to image classification, facial recognition and computer vision \cite{kuznetsova2016exploiting, liao2018semantic, luo2017zero, mettes2017spatial, naha2016object, shu2015weakly, song2022afusion, tang2016generalized, wang2015zero, xu2017transductive, yang2015zeroshot, zhang2016zero}. This happens to such a high degree that researchers usually refer to just these applications as zero-shot \cite{luo2017zero, yang2015zeroshot}, despite the emerging existence of some other classification applications \cite{levy2017zero, robyns2017physical, shigeto2015ridge, wang2017zero}. Reviewing the state-of-the-art of the zero-shot approaches, one can find that they are mostly proposed for classification task (predicting a label) rather for regression task (predicting a continuous value), as it is our case of predicting pollutant concentrations. The main drawback of these approaches is that they adopt decisions and establish strategies that are specific of a classification task. Hence, extending or adapting them to a regression task is non-obvious, non-direct and even unfeasible \cite{reis2018hyperprocess}. The so-called regression-based zero-shot methods \cite{liao2018semantic, luo2017zero, palatucci2009zero, shigeto2015ridge, yang2015zeroshot} have paved the way among the existing strategies to cope with zero-shot learning. Despite these approaches being labelled as regression-based methods, they do not solve regression zero-shot learning; otherwise, they cope with classification zero-shot. But they are referred as regression-based zero-shot methods because they commonly include regression learning to make projections to map instances in certain spaces depending on the classes of the classification, typically feature or side information spaces, as a previous step of performing the classification into classes. So far, very few approaches are found in the literature for zero-shot regression that would solve several other problems beyond the pollutant concentration prediction. For instance, in an industrial environment \cite{morariu2020machine}, predicting the performance of the parts of a machine given different system’s states, where the parts are interchangeable and have certain specifications or characteristics known beforehand, what it can be extended to unknown and unobservable parts for unknown and unobservable system's states. Also, in agricultural production \cite{belcore2021precision}, static environment (soil moisture, position characteristics, weather...) and human defined factors (irrigation, fertilization, pest control). There are lots of research on predicting the agricultural production from the human defined factors, but these predictions are based on just certain static environment. Considering different static environments, or even, new locations for them, predicting the agricultural production becomes a zero-shot regression task. Another application could also be for monitoring, control and modeling of water treatment for human, industrial, or agricultural water consumption \cite{hu2023theutility}. The ability to incorporate side information derived from the environment opens up a promising range of applications for water treatment prediction and control, such as pollution control, water quality prediction, and water quality prediction. The pioneers of providing zero-shot methods for regression are two preliminary works \cite{reis2018hyperprocess, zhang2020cazsl} not able to cope with a general-purpose regression zero-shot task \cite{fdez2022target}. The former \cite{reis2018hyperprocess} reduces the experiments to a single toy example based on a beta distribution with just two features for the side information and one common instance feature. In fact, if one tries to perform experiments over datasets with higher dimensions in features and side information, the available software reports the message "expected 1D vector for $x$". The latter \cite{zhang2020cazsl} predicts the future position of a piece that is pushed by a robotic arm, given the present location through deep learning approach specifically built for this purpose, avoiding to be used as a general purpose zero-shot regressor. Quite recently, another work \cite{fdez2022target} overcomes this issue and has presented two novel approaches with appealing performance in a general-purpose zero-shot regression framework. Both approaches treat side information different from common features \cite{mollaysa2019learning} and provide unobserved target models through a two-stage procedure. They coincide in providing observed target models in the first stage, whereas they differ in the way side information is taken in the second stage for providing unobserved target models. One of them is a simple relationship approach inspired by the inverse distance weighting. Particularly, predictions using the observed target models are weighted by the similarity of each observed target with the unobserved target in order to provide unobserved target predictions. The main disadvantage of this approach is that side information is just considered a posteriori in the testing stage of the second phase. In addition, it just interpolates predictions of the observed target models, whose values will be bounded in a certain range, then its generalization power is quite limited. The other method arose in an attempt of overcoming these drawbacks. The result was a correspondence method that takes the side information into the learning process of the second phase in order to increase the generalization power of the predictions for the unobserved targets. Particularly, the method learns the parameters of the unobserved target models from both the observed target models and the side information. This alternative actually improves the predictive performance of the unobserved targets. However, an assumption of a linear relationship between features and targets must be established. The main disadvantage of these two approaches is that they deal with side information (information of the surroundings of the meteorological stations) separately (in different phases) from the common features (weather conditions), avoiding a global optimization. At this point, one can notice that, on the one hand, taking side information as common features directly handles zero-shot regression in just one learning process obtaining a global optimization, but it does not provide promising performance \cite{fdez2022target} because the same treatment of both kinds of information is not a good practice. On the other hand, exploiting side information in a more strategic and specific way taking it separately from the common features, although, so far, into separate phases, has been shown to improve the predictive performance \cite{fdez2022target}, but it does not provide a global optimization. This situation sheds the light of tackling zero-shot regression directly in just one learning process, but adequately handling features and side information, each one differently and according to their nature. The contribution of this paper goes in this direction and the proposal consists of a novel one-stage learning approach for zero-shot regression based on a kernel definition that properly integrates both features and side information in the same learning process. The proposed approach experimentally exhibits its superiority in performance with regard to other existing approaches for zero-shot regression, one of them consisting of treating side information as common features being the other the two the above-mentioned relationship method and correspondence method.

The rest of the paper is organized as follows. Section \ref{sec:related} describes some related work. Section \ref{sec:zeroshot} details the zero-shot regression statement and the state-of-the-art methods available in the literature. Then, the new proposal consisting of a one-stage learning process that jointly integrates side information and common features is detailed in Section \ref{sec:proposal}. In Section \ref{sec:experiments} the description of the experiments and the discussion of the results are exposed. Finally, Section \ref{sec:conclusions} draws some conclusions and proposes some lines of research for future work.

\section{Related Work} \label{sec:related}
Zero-shot scenario opens the possibility of supposing an inductive rather than transductive learning paradigm with regard to targets \cite{wang2019asurvey}. Target inductive learning induces models for generic unknown and unobservable targets, whereas target transductive learning induces models for specific unknown and unobservable targets. Precisely, the own goal that characterizes the zero-shot scenario of inducing models for targets for which instances are not available (unobservable targets) makes possible contemplate both scenarios with regard to the targets. Notice that the side information of both observed and unobserved targets is supposed to be available. Hence, the difference between target transductive and target inductive learning is in fact reflected in whether the side information of the unobserved targets is including in the training phase of unobserved target models (transductive) or not (inductive).  This situation does not happen in classical machine learning, for which models are required for just targets with available instances (observable targets). The majority of the works about zero-shot, including ours, assumes inductive learning for targets, but there are some that deal with target transductive learning \cite{rahman2019transductive}. 

The approaches to cope with zero-shot task can be split into instance-based or model-based methods \cite{fdez2022target, wang2019asurvey} independently if they are for a classification task (predicting a label) or a regression task (predicting a continuous value). The former adopts diverse strategies such as extraction or learning in order to provide instances for the unobserved targets, from which unobserved target models are learned afterwards, whereas the latter the unobserved targets are directly learned from the information available. The model-based approaches, in which fall our proposal, can be split into relationship, correspondence and combination methods depending on how they manage to provide unobserved target models \cite{wang2019asurvey}. On one hand, both relationship and correspondence approaches learn observed target models. But they differ in the way they exploit the side information. Relationship assumes the existence of a relationship function between observed and unobserved targets. This function together with the observed target models are taken to obtain unobserved target models. However, correspondence methods learn the correspondence between the observed target models and the observed target side information. On the other hand, combination methods decompose the observed and unobserved targets into basic elements, learn a model per each basic element and finally they combine the basic element models via an inference process in order to obtain the unobserved target models. 

As commented before, classification into a set of finite classes has extensively been the focus of researches in the zero-shot learning, leaving a lack of approaches to cope with zero-shot regression (predicting continuous values as they are the concentration of the pollutants). In general, classical classification methods has been adaptable and extensible to classical regression or at least it was possible to transfer the ideas established for the classical classification to the classical regression. However, in the context of zero-shot learning, the ideas taken for zero-shot classification are highly influenced by the classes of the classification and not easily extensible for zero-shot regression, even they are unfeasible to transfer. Great efforts have been made to provide a fan of approaches for zero-shot classification overcoming the lack of instances and exploiting the side information. Particularly, the so-called generalized zero-shot classification methods \cite{li2022micro, pourpanah2022areview} have captured the attention of the research in the last years. They arise in an attempt of fill the existing gap of the traditional zero-shot classification methods, which have limited generalization power to the unobservable classes, since they combine semantic (side) information with common features of just the observable classes. Then, the models they induced tend to classify instances of the unobservable classes as belonging to one of the observable classes. Generalized zero-shot classification methods can be split into embedding-based methods, generative-based methods and common space methods. Embedding-based methods learn an embedding space to relate the common (visual in image classification) features of observable classes with their corresponding semantic information (side information) \cite{chen2022transzero, chen2022mutually, huynh2020finegrained}. They learn a projection function able to recognize unobservable classes by measuring the similarity level between the semantic (side) information of the observable and unobservable classes in the embedding space.  These methods are biased to the observable classes, so their generalization power to unobservable classes is limited. Generative-based methods \cite{chen2021free, narayan2020latent, xian2019fvaegan} learn a model to generate instances (images) or common (visual) features for the unobservable classes based on the instances of observable classes and semantic (side) information of both kind of classes. By generating instances for unobservable classes, the task is converted into a classical classification task. In one sense, these methods overcome the bias problem of the embedding-based methods, since the models are learned to be able to classify instances from both observable and unobservable classes. Finally, common-space methods \cite{chen2021HSVA, schonfeld2019generalized} learn a common representation space into which both common (visual) features and semantic (side) information are projected in order to get an effective knowledge transfer. Latent features are built, which are the ones that contains the whole information coming from the unobservable classes. 

Similar frameworks to zero-shot learning are few-shot learning \cite{kang2019few,peng2019fewshot, song2022afusion} and one-shot learning \cite{koch2015siamese,feifei2006oneshot}. In both scenarios, unobservable targets have some instances available, but a reduced and a limited number of them. The difference lies in that in the former few instances are available whereas in the latter just one instance is available. 


Transfer learning \cite{farahani2020concise, zhuang2020comprehensive} is a field closely related to zero-shot learning. In transfer learning, the source domain and source task respectively are the observable instances and targets. The counterpart of the unobservable instances and targets are called the target domain and target task. Hence, the aim of transfer learning consists of extracting knowledge from the source domain and task and transferring it to the target domain in order to cope with the target task. Transfer learning also includes inductive and transductive paradigms, but adds the unsupervised option \cite{pan2010transferlearning}; all defined in terms of properties that the source and target domains and tasks satisfy. 
Under an inductive paradigm, source and target tasks differ, no matter if the respective domains coincide or not. Under a transductive setting, source and target domains differ, whereas the source and target tasks coincide. Finally, under an unsupervised scenario, the target and source tasks differ as it happens under an inductive paradigm, but the tasks fall into unsupervised learning. In the context of this paper, the source and target task differ, then leading to an inductive transfer learning paradigm. Besides, the source and target domains coincide, since both deal with the same features. More classical transfer learning required the availability of some instances in the target domain whatever inductive or transductive paradigms. Only more recent transfer learning approaches \cite{campagna2020zero, socher2013zero, yang2019zero} cope with situations deprived of instances in the target domain, but unfortunately, they are designed exclusively for classification, which is not easily adaptable to regression. Besides, they are customized for what they were designed, for instance, for outlier detection \cite{socher2013zero}, extracting specific image features \cite{yang2019zero}, or obtaining synthesized dialogue instances \cite{campagna2020zero}.  Hence, they are not applicable to a general-purpose zero-shot regression task.

\section{Zero-shot regression statement and state-of-the-art methods} \label{sec:zeroshot}
This section formally states the zero-shot regression task and also formally discusses the strategies followed by the state-of-the-art methods. The formal statement of the zero-shot regression task will be defined in terms of inductive learning, since, as commented in Section \ref{sec:related}, our assumption is that unobserved targets are supposed to be unknown and generic. Let $\mathcal{X}$, $\mathcal{S}$ and $\mathcal{Y}$ respectively denote the feature space of instances, the feature space of targets (side information) and the image space of the predictions. Hence,
\begin{itemize}
\item Let $\mathcal{T}^{o}=\{t_{i}^{o}\in \mathcal{S} \}_{i=1}^{m_o}$ be the set of observed targets, where $m_o$ is the number of them. The notation $t_{i}^{o}\in \mathcal{S}$ represents the side information of the observed targets, that will be in form of features, as commented in Section \ref{Introduction}. Let $t^{u}\in \mathcal{S}$ be the side information (feature representation) of a generic unobserved target such that $t^{u}\notin \mathcal{T}^{o}$.
\item Let $\mathcal{D}^{o}=\{(x_{j}^{o},y_{j}^{o})\in  \mathcal{X}\times\mathcal{Y}\}_{j=1}^{n_{o}}$ be the set of instances for the observed targets, where $n_o$ is the number of them. Then, $\mathcal{D}^{o,t_i^o}\subset \mathcal{D}^{o}$ will denote the set of instances of the observed target $t_i^o\in \mathcal{T}^{o}$ for all $i=1,\dots, m_o$. Let $x^u$ be an instance of the unobserved target $t^u\notin \mathcal{T}^{o}$ whose prediction is $y^u\in \mathcal{Y}$ such that $(x^u, yu)\notin \mathcal{D}^{o}$.
\end{itemize}

Therefore, the inductive zero-shot regression task consists of learning a function $f:\mathcal{X}\times \mathcal{S}\rightarrow \mathcal{Y}$ from $\mathcal{D}^{o}$ and $\mathcal{T}^{o}$ able to predict $y^{u} \in \mathcal{Y}$ for a generic instance $x^u \in \mathcal{X}$ of a generic unobserved target $t^u\in \mathcal{S}$ (in the transductive scenario the function to learn would be $f_{t^u}:\mathcal{X}\rightarrow \mathcal{Y}$ from $\mathcal{D}^{o}$, $\mathcal{T}^{o}$ and an specific unobserved target $t^u$   able to predict $y^{u} \in \mathcal{Y}$ for a generic instance $x^u \in \mathcal{X}$, see \cite{fdez2022target} for more details).
The next subsections formally and briefly detail the state-of-the-art methods for zero-shot regression.

\begin{figure}
	\centering
	\subfigure[\base]{\includegraphics[width=8cm, trim={4.6cm 7.5cm 4.1cm 4.7cm}, clip]{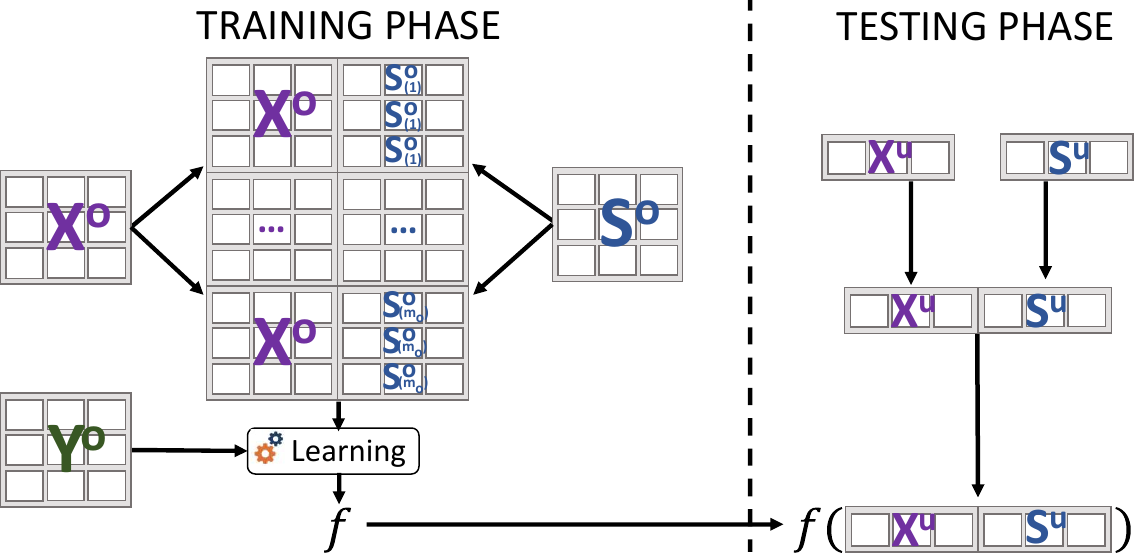}\label{fig:Schemaa}}
	\subfigure[\sr]{\includegraphics[width=11cm, trim={0 5cm 3.4cm 3cm}, clip]{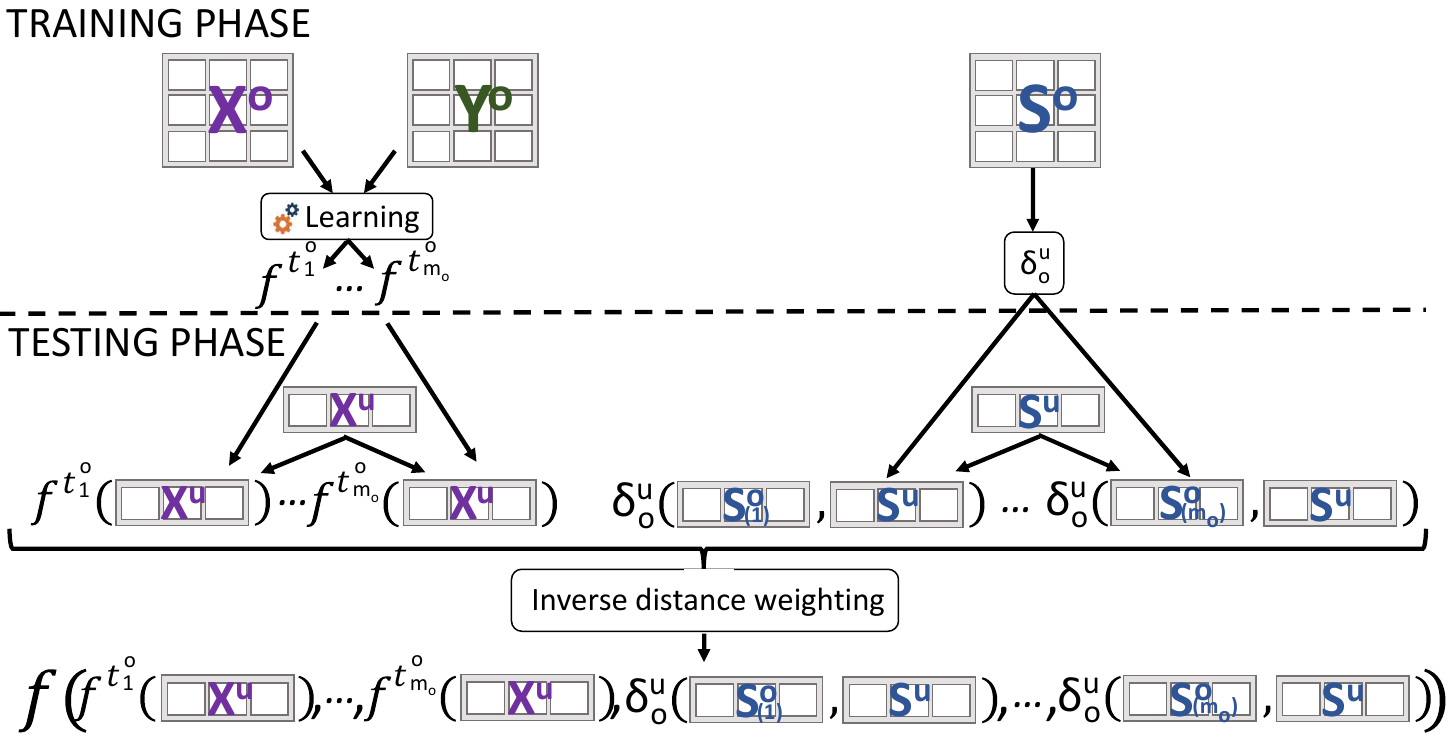}\label{fig:Schemab}}
	\subfigure[\mplc]{\includegraphics[width=10cm, trim={0 6.2cm 4cm 0.5cm}, clip]{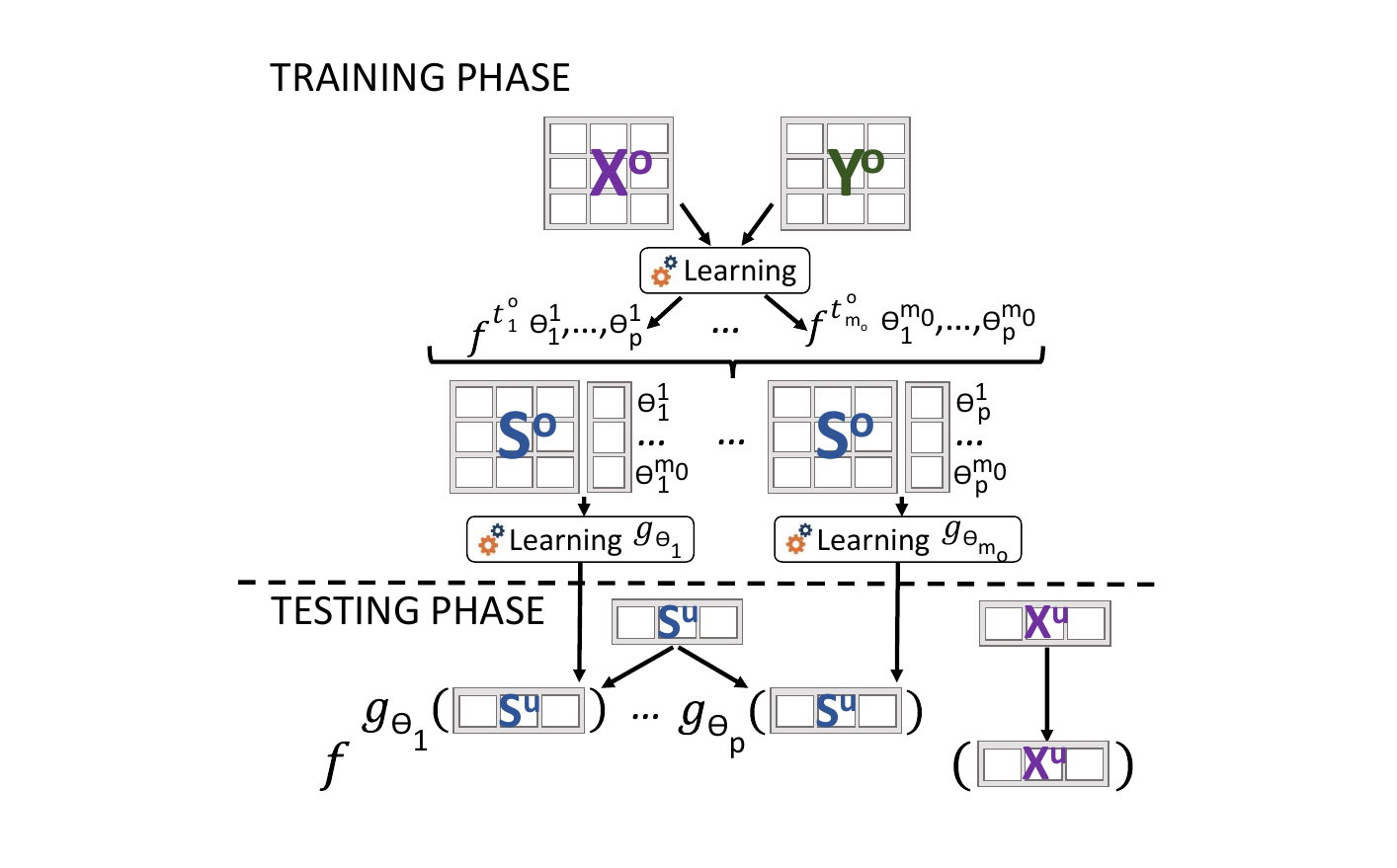}\label{fig:Schemac}}
	\caption{Baseline, similarity relationship and model parameter learning correspondence methods for zero-shot regression} \label{fig:Schemas}
\end{figure}

\subsection{Baseline method} \label{sec:base}
The baseline method (\base{}) \cite{fdez2022target} was designed to be a point of reference in an attempt of avoiding ignoring side information. The side information is included as convectional features. Hence, the instances of the same target will have the same values in these features (see Figure \ref{fig:Schemaa}). In the training phase, just one learning process takes place. The features are built taking $\mathcal{D}^{o}_{\mathcal{X}}=\{x_{j}^{o} \in \mathcal{X} \}_{j=1}^{n_{o}}$ (matrix $X^o$) and $\mathcal{T}^{o}=\{t_{i}^{o}\in \mathcal{S} \}_{i=1}^{m_o}$ (matrix $S^o$) and joining (concatenating) each target side information $t_i^o\in\mathcal{T}^{o}$ to their correspondent instances $\mathcal{D}^{o, t_i^o}_{\mathcal{X}}\subset \mathcal{D}^{o}_{\mathcal{X}}$. The goal consists of taking the correspondent prediction values of $\mathcal{D}^{o}_{\mathcal{Y}}=\{y_{j}^{o} \in \mathcal{Y}\}_{j=1}^{n_{o}}$ (matrix $Y^o$), again taking into account the correspondent predictions $\mathcal{D}^{o, t_i^o}_{\mathcal{Y}}\subset \mathcal{D}^{o}_{\mathcal{Y}}$ of each target $t_i^o\in \mathcal{T}^{o}$ for all $i=1,\dots, m_o$. Then, a function $f:\mathcal{X}\times \mathcal{S}\rightarrow \mathcal{Y}$ is induced. In the testing phase, the features of an unobserved instance $x^{u}$ (vector $X^u$) of an unobserved target $t^u$ are joined (concatenated) to the side information of the unobserved target $t^u$ (vector $S^u$) to feed $f$ and provide the prediction $f(x^u,t^u)$. This method works under the hypothesis of existing an equal kind of relationship between features and targets and between this relationship and the side information, since common features and side information are treated equally. However, this assumption is quite limited and tied, since it is not usually satisfied. In fact, the nature of side information is different from the common features. 

\subsection{Similarity relationship method} \label{sec:sr}
The similarity relationship method (\sr{}) \cite{fdez2022target} is an appealing method to cope with zero-shot regression due to its simplicity. The main point of this method is the establishment of a relationship function $\delta^{o,u}$ between observed and unobserved targets. Particularly, $\delta^{o,u}$ establishes the similarity between targets in terms of the inverse of a distance, that is, given a distance $d$, $\delta^{o,u}$ is defined as
$$ \begin{array}{lcll}
\delta^{o,u}:&\mathcal{T}^o\times \mathcal{T}^u&\rightarrow &\mathbb R\\
&(t^o,t^u)&\rightarrow &1/d(t^o,t^u)
\end{array}$$
The method works in two stages (see Figure \ref{fig:Schemab}), namely:
\begin{itemize}
\item [i)]The first stage takes place in the training phase and consists of learning the observed target models $f^o=\{f^{t^o_i}\}_{i=1}^{m_o}$ from instance features $\mathcal{D}^{o}_{\mathcal{X}}=\{x_{j}^{o} \in \mathcal{X} \}_{j=1}^{n_{o}}$ (matrix $X^o$)  and prediction values $\mathcal{D}^{o}_{\mathcal{Y}}=\{y_{j}^{o} \in \mathcal{Y}\}_{j=1}^{n_{o}}$ (matrix $Y^o$), then, ignoring side information $\mathcal{T}^{o}=\{t_{i}^{o}\in \mathcal{S} \}_{i=1}^{m_o}$ (matrix $S^o$). Each $f^{t^o_i}$ has been induced taken the correspondent $\mathcal{D}^{o, t_i^o}_{\mathcal{X}}$ and $\mathcal{D}^{o, t_i^o}_{\mathcal{Y}}$ that concerns to the observed target $t^o_i$, that is, taking the subset of rows of matrix $X^o$ of the features and the subset of rows of matrix $Y^o$ of the predictions that correspond to target $t^o_i$. 
\item [ii)] The second stage happens in the testing phase and it is the stage in which side information $\mathcal{T}^{o}=\{t_{i}^{o}\in \mathcal{S} \}_{i=1}^{m_o}$ (matrix $S^o$) comes into play. Then, an unobserved instance $x^{u}$ (vector $X^u$) of an unobserved target $t^u$ (vector $S^u$) is evaluated using the models induced in the first stage $f^o=\{f^{t^o_i}\}_{i=1}^{m_o}$. Then, the set of prediction values $f^o(x^u)=\{f^{t^o_i}(x^u)\}_{i=1}^{m_o}$ are obtained. Finally, these predictions are aggregated through a normalized weighting procedure induced by the correspondent similarity $\{\delta^{o,u}(t_i^o, t^u)\}_{i=1}^{m_o}$ of the observed targets $\mathcal{T}^{o}=\{t_{i}^{o}\in \mathcal{S} \}_{i=1}^{m_o}$ (matrix $S^o$) and the unobserved target $t^u$ (matrix $S^u$) in order to produce the prediction $f(x^u, t^u)$ for the unobserved instance $x^u$ of the unobserved target $t^u$. Hence, $f:\mathcal{X}\times \mathcal{S}\rightarrow \mathcal{Y}$ is defined as
$$f(x^u, t^u)=\frac{1}{\sum_{i=1}^{m_o}{\delta^{o,u}(t_i^o,t^u)}}\cdot \sum_{i=1}^{m_o}\delta^{o,u}(t_i^o,t^u)\cdot f^{t^o_i}(x^u)$$
\end{itemize}

The similarity function based on the inverse of a distance allows guaranteeing that the models of the least similar observed targets to the unobserved target make less influence on the prediction of the unobserved instance than the models of the most similar observed targets.
This method has shown to perform well in spite of its simplicity. However, the generalization power may be compromised because the method only interpolates values of the observed target models to provide predictions for the unobserved instances of unobserved targets. Besides, the side information is not included in the learning process; otherwise, it is exploited in the testing phase. Hence, instance features and side information are exploited separately in two different phases. 

\subsection{Model parameter learning correspondence method} \label{sec:mplc}
Model parameter learning correspondence method (\mplc{})  \cite{fdez2022target} arose in an attempt of improving the generalization power of \sr, including the side information in a learning process. Particularly, the approach tries to learn the correspondence between the observed targets and the correspondent observed targets model parameters, which may potentially increase the generalization power of the unobserved target models (see Figure \ref{fig:Schemac}). The method also works in two stages, as \sr{}. However, in this case, both stages take place in the training phase, unlike \sr{}, whose second phase takes place in the testing phase. The stages of this approach are the following:
\begin{itemize}
\item [i)] The first stage of \mplc\ coincides to the first stage of \sr. Then, observed target models $f^o=\{f^{t^o_i}\}_{i=1}^{m_o}$ from instance features $\mathcal{D}^{o}_{\mathcal{X}}=\{x_{j}^{o} \in \mathcal{X} \}_{j=1}^{n_{o}}$ (matrix $X^o$)  and prediction values $\mathcal{D}^{o}_{\mathcal{Y}}=\{y_{j}^{o} \in \mathcal{Y}\}_{j=1}^{n_{o}}$ (matrix $Y^o$), but ignoring side information $\mathcal{T}^{o}=\{t_{i}^{o}\in \mathcal{S} \}_{i=1}^{m_o}$ (matrix $S^o$) are induced in the first stage. Again, each $f^{t^o_i}$ has been induced taken the correspondent $\mathcal{D}^{o, t_i^o}_{\mathcal{X}}$ and $\mathcal{D}^{o, t_i^o}_{\mathcal{Y}}$ that concerns to the observed target $t^o_i$, that is, taking the subset of rows of matrix $X^o$ of the features and the subset of rows of matrix $Y^o$ of the predictions that correspond to target $t^o_i$. 
\item[i)] The second stage of \mplc\ differs from the second stage of \sr. In the case of this method (\mplc), another learning procedure takes place in the second stage. In this case, the aim is to learn the parameters of the unobserved target models. The parameters $\Theta=\{(\theta_1^i,\cdots, \theta_p^i)\}_{i=1}^{m_o}$ of the observed target models $f^o=\{f^{t^o_i}\}_{i=1}^{m_o}$ resume the relationship between features and targets for the observed targets. This parameters are taken together with the side information $\mathcal{T}^{o}=\{t_{i}^{o}\in \mathcal{S} \}_{i=1}^{m_o}$ (matrix $S^o$) of the observed targets in the learning process. More in detail, a set of $p$ (as many as number of model parameters) learning procedures are carried out, whose features are the observed target side information $\mathcal{T}^{o}=\{t_{i}^{o}\in \mathcal{S} \}_{i=1}^{m_o}$ (matrix $S^o$) for all them and whose targets are respectively the $\Theta_1=\{\theta_1^i\}_{i=1}^{m_o}$, $\dots$, $\Theta_p=\{\theta_p^i\}_{i=1}^{m_o}$ of the observed target models $f^o=\{f^{t^o_i}\}_{i=1}^{m_o}$. Then, the $p$ learning processes induces the $g_\theta=\{g_{\theta_j}\}_{j=1}^p$ set of models defined over $\mathcal{S}$. In the testing phase, the side information of an unobserved target $t^u$ (matrix $S^u$) feeds all the $g_\theta=\{g_{\theta_j}\}_{j=1}^p$ to provide the parameters $\{\theta_j^u\}_{j=1}^p$ for the model of the unobserved target $t^u$. Finally, the function $f:\mathcal{X}\times \mathcal{S}\rightarrow \mathcal{Y}$  is configured from these parameter value predictions $g_\theta(t^u)=\{\theta_j^u=g_{\theta_j}(t^u)\}_{j=1}^p$ leading to a function $f^{g_\theta(t^u)}:\mathcal{X} \rightarrow \mathcal{Y}$ that gets ready to evaluate any unobserved instance $x^u$ (matrix $X^u$) of the unobserved target $t^u$. 
\end{itemize}
The main disadvantage of this method is that is necessary to assume a linear relationship (see \cite{fdez2022target} for details) in order to state the learning procedure of the second phase and this assumption do not hold in general.
 
\section{Direct side information learning for zero-shot regression} \label{sec:proposal}
Despite \mplc{} includes the target side information into a learning process in order to provide more generalization power than \sr, it does it in a separate stage different from the observed target model learning process. Hence, learning processes are locally optimized. The proposal of this paper unifies both instance and target (side information) features (all the information available) in a one-stage learning process, then, obtaining a globally optimized learning process, as \base{} method does it. However, and meanwhile \base{} method treats equally feature instance description and side information, our novel approach adequately integrates features and side information thought a kernel definition according to the nature of both kinds of information. Therefore, the function $f:\mathcal{X}\times \mathcal{S}\rightarrow \mathcal{Y}$ is directly learned from both instance features and side information. If $a_x$ and $a_s$ respectively are the instance feature and side information sizes, $x=(x_1,\cdots, x_{a_x})$ is a feature instance description and $s=(s_1,\cdots, s_{a_s})$ is a target side information description, let us restrict our proposal to the linear case, i) in the relationship between features and predictions and ii) in the relationship between the side information and the relationship of i) (a non-linear scenario will be proposed as future work). Then, $f(x,s)$ adopts the following form
\begin{equation}\label{eq:1}
\begin{array}{lcll}
f:&\mathcal{X}\times \mathcal{S}&\rightarrow & \mathcal{Y}^u\\
&(x,s)&\rightarrow & f(x,s)=f_{\beta}(s)+
\sum_{i=1}^{a_x} f_i(s)\cdot x_i
\end{array}
\end{equation}
where $f_\beta(s)$ and each $f_i(s)$ are in turn linear functions of $s$, that is,
\begin{equation}\label{eq:2}
 f_i(s)=\beta_{i}+\sum_{j=1}^{a_s}\alpha_{i,j} \cdot s_{j}  \qquad i=1,\cdots, a_{x}
\end{equation}
\begin{equation}\label{eq:3}
 f_\beta(s)=\beta_{\beta}+ \sum_{j=1}^{a_s}\alpha_{\beta,j} \cdot s_{j}
\end{equation}
where $\{\beta_i\}_{i=1}^{a_x}$, $\{\alpha_{i,j}\}_{i,j=1,1}^{a_x,a_s}$, $\beta_\beta$ and $\{\alpha_{\beta,j}\}_{j=1}^{a_s}$ are the parameters of the linear functions. Then, including the expressions of the Equations (\ref{eq:2}) and (\ref{eq:3}) in the Equation (\ref{eq:1}), the expression of $f(x,s)$ will be
\begin{equation}\label{eq:4}
f(x,s)=\left (\beta_{\beta}+ \sum_{j=1}^{a_s}\alpha_{\beta,j} \cdot s_{j}\right)+ \sum_{i=1}^{a_x} \left(\beta_{i}+\sum_{j=1}^{a_s}\alpha_{i,j} \cdot s_{j}\right)\cdot x_i
\end{equation}
or equivalently 
\begin{equation}\label{eq:5}
f(x,s)=\beta_{\beta}+ \sum_{j=1}^{a_s}\alpha_{\beta,j} \cdot s_{j}+ \sum_{i=1}^{a_x} \beta_{i}\cdot x_i+ \sum_{i=1}^{a_x} \sum_{j=1}^{a_s}\alpha_{i,j} \cdot s_{j}\cdot x_i
\end{equation}

At this point, we will define a mapping function $\phi$ from $\mathcal{X}\times \mathcal{S}$ space into a Hilbert space $\mathcal{H}$ and a linear function $g$ defined over the image space $\mathcal{H}$ of $\phi$ such that the function $f$ will be expressed as
\begin{equation}\label{eq:6}
f(x,s)=g(\phi(x,s))
\end{equation}

For this purpose, the mapping function $\phi$ will be
\begin{equation}\label{eq:7}
\begin{array}{lcll}
\phi:&\mathcal{X}\times \mathcal{S}&\rightarrow &\mathcal{H}\\
&(x,s)&\rightarrow &\phi(x,s)=((1,x)^T\otimes_{K}(1,s)^T)^T
\end{array}
\end{equation}
where $\otimes_{K}$ denotes the Kronecker product of vectors. The Kronecker product of vectors, also called matrix direct product, means to vectorize the outer product of vectors, denoted by $\otimes_{O}$. Then, the Equation (\ref{eq:7}) can be expressed in the following way
\begin{equation}\label{eq:8}
\phi(x,s)=\left(vec\left ((1,x)^T\otimes_{O}(1,s)^T\right )\right )^T \footnote{$vec$ is the function that vectorizes a matrix, that is, it converts a matrix into a column vector by concatenating the columns.}. 
\end{equation}
Also, the outer product of vectors means to multiply each element of a vector by each element of the other vector. Then, the Equation (\ref{eq:8}) can be expressed as
\begin{equation}\label{eq:9}
\phi(x,s)=\left (vec\left ((1,x)^T (1,s)\right )\right )^T. 
\end{equation}
Expanding the Equation (\ref{eq:9}) leads to the following expression of $\phi$ 
\begin{multline}\label{eq:10}
\phi(x,s)=\left (vec  \left (\begin{bmatrix}
1\\
x_1\\
\vdots\\
x_{a_x}\\
\end{bmatrix} 
\begin{bmatrix}
1 & s_1 & \cdots & s_{a_s}\\
\end{bmatrix} \right)\right)^T= \\
 =\left (vec  \left( \begin{bmatrix}
1 & s_1 & \cdots & s_{a_s}\\
x_1 & x_1\cdot s_1 & \cdots & x_1\cdot s_{a_s}\\
\vdots & \vdots & \ddots & \vdots\\
x_{a_x} & x_{a_x}\cdot s_1 & \cdots & x_{a_x}\cdot s_{a_s}\\
\end{bmatrix}\right)\right)^T
\end{multline}
and, finally, performing the $vec$ function and the transpose operator afterwards in the Equation (\ref{eq:10}), $\phi$ will be

\begin{multline}\label{eq:11}
\phi(x,s)=\left (\begin{matrix} 1, & x_1, & \cdots, & x_{a_x}, \end{matrix} \right.\\
\left.\begin{matrix}  s_1, & x_1\cdot s_1,& \cdots, & x_{a_x}\cdot s_1,&   \end{matrix} \right.\\
\left.\begin{matrix} \cdots, \cdots, \cdots, \cdots\end{matrix} \right.\\
\left. \begin{matrix} & s_{a_s},& x_1\cdot s_{a_s},&\cdots,& x_{a_x}\cdot s_{a_s} \end{matrix}\right)
\end{multline}

Let us notice that the components of $\phi$ are the unit and the components of $x$, that is, $(1,  x_1,  \cdots,  x_{a_x})$, concatenated by these same components multiplied by the first component of $s$, that is, $(s_1,  x_1\cdot s_1, \cdots,  x_{a_x}\cdot s_1)$ and so on until the same components multiplied by the last component of $s$, that is, $(s_{a_s}, x_1\cdot s_{a_s},\cdots, x_{a_x}\cdot s_{a_s})$. Then, the components of $\phi$ are all the monomials of a $2-$degree polynomial, except two kinds of monomials, namely, i) the squared ones of the kind $x_i^2$ and $s_j^2$ and ii) the ones of the kind $x_i\cdot x_j$ and $s_i\cdot s_j$ with $i\neq j$. 
 
Then, the linear function $g$ will be defined as a linear combination of the monomial components of $\phi(x,s)$, that is,
\begin{equation}\label{eq:12}
g(\phi(x,s))=\beta_{\beta}\cdot 1+ \sum_{i=1}^{a_x} \beta_{i}\cdot x_i+  \sum_{j=1}^{a_s} \left (\alpha_{\beta,j} \cdot s_{j} +\sum_{i=1}^{a_x} \alpha_{i,j} \cdot x_{i}\cdot s_j \right)
\end{equation}

\begin{figure}[H]
	\centering
	\includegraphics[width=13.5cm, trim={1.5cm 5cm 4cm 4.5cm}, clip]{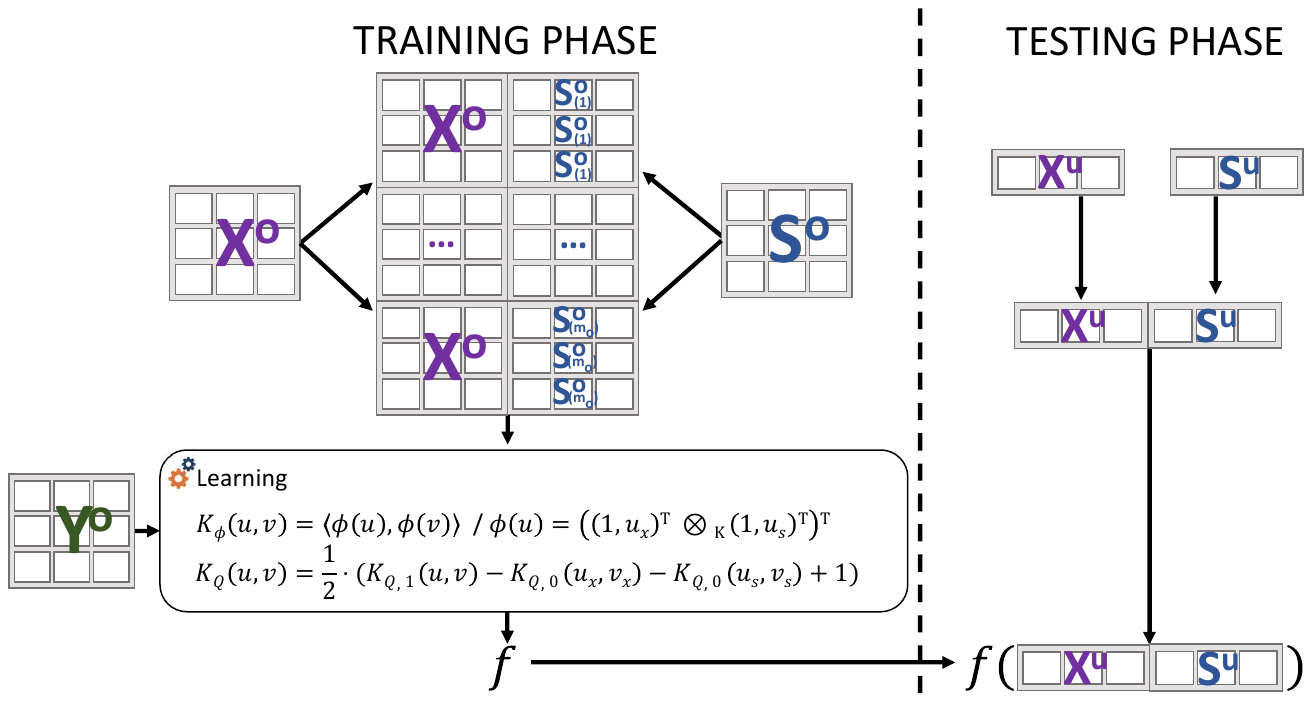}
	\caption{Training and testing phases for the DSIL method}
	\label{fig:schemaGlobal}
	\centering
\end{figure}

Finally, let us notice that reordering the terms of the Equation (\ref{eq:12}), one can easily obtain the expression of $f(x,s)$ in the Equation (\ref{eq:5}). Therefore, $f(x,s)$ can be expressed as a linear function $g$ defined over the image space of $\phi$, as stated in the Equation (\ref{eq:6}). Now, let us propose three different ways of inducing the function $f$. The first one will consist of applying the mapping function $\phi$ and inducing the linear function $g$ afterwards according to the Equation (\ref{eq:6}). The second and third ways define a kernel $K$ that computes the inner product in the image space of $\phi$. However, the second way explicitly applies the function mapping $\phi$, whereas the third way straightly computes the inner product in the image space and avoids applying the mapping function. The result is a one-phase learning method that directly includes the side information in this learning process. This is the reason why the method is called Direct Side Information Learning (\dsil). Figure \ref{fig:schemaGlobal} displays the training and testing phases for this new approach. The structure is analogous to that of the \base\ approach (see Figure \ref{fig:Schemaa}), just changing the linear kernel by the new proposed kernels. Next subsections \ref{sec:kernelphi} and \ref{sec:kernelq} deal with the second and third ways of inducing the function $f$, which both involve a kernel definition. Also, subsection \ref{sec:toyexample} describes a toy example in order to illustrate the way DSIL works. Finally, subsection \ref{sec:complexity} analices the three ways of inducing the function $f$ in terms of computational cost.

\subsection{A kernel definition using the mapping function $\phi$} \label{sec:kernelphi}
This section proposes a kernel associated with the $\phi$ mapping function in order to establish a gangway from linearity to non-linearity in terms of the inner dot product. The kernel definition allows performing the mapping and the inner product simultaneously. In this case, the kernel can be defined in terms of the linear kernel and $\phi$ using one of the closure properties of the kernels \cite{shawe2004kernel} as


\begin{equation}\label{eq:13}
K\left(\left(x^{(1)},s^{(1)}\right),\left(x^{(2)},s^{(2)}\right)\right)=\langle \phi \left(x^{(1)},s^{(1)}\right), \phi \left(x^{(2)},s^{(2)}\right) \rangle
\end{equation}

Using the expression of the mapping function $\phi$ exposed in the Equation (\ref{eq:8}) and dispensing with the $vec$ function, the kernel can be expressed in terms of the outer product of vectors and the Hadamard product (also called the element-wise product, entry wise product or Schur product) of a matrix, which is denoted by $\odot$ and consists of computing the product of the matrixes element-by-element. 
\begin{multline}\label{eq:14}
K\left(\left(x^{(1)},s^{(1)}\right),\left(x^{(2)},s^{(2)}\right)\right)=\sum \left (\left (1, x^{(1)}\right)^T \otimes_{O} \left(1, s^{(1)}\right)^T\right)\odot \\ 
\left(\left(1, x^{(2)}\right)^T \otimes_{O} \left(1, s^{(2)}\right)^T\right)
\end{multline}

Computing the outer and Hadamard products and the summation of the elements of the resultant matrix in the Equation (\ref{eq:14}), the expression of $K\left(\left(x^{(1)},s^{(1)}\right),\left(x^{(2)},s^{(2)}\right)\right)$ becomes

\begin{multline}\label{eq:15}
K\left(\left(x^{(1)},s^{(1)}\right),\left(x^{(2)},s^{(2)}\right)\right)=1+\sum_{i=1}^{a_x} x_i^{(1)}x_i^{(2)}+\sum_{i=1}^{a_s} s_i^{(1)}s_i^{(2)}+\\
+\sum_{i=1}^{a_s}\sum_{j=1}^{a_x} \left (x_j^{(1)}s_i^{(1)}\right)\left (x_j^{(2)}s_i^{(2)}\right)
\end{multline}

The main disadvantage of this kernel definition is precisely the necessity of computing the $\phi$ mapping, since the expression of the Equation (\ref{eq:15}) must be computed for all pairs $(x,s)$ and in addition it is of quadratic order with regard to the features. Hence, an alternative will be to define the kernel thought an expression able to compute the inner product in the image space of $\phi$ but avoiding the use of the $\phi$ mapping and performing a linear order computation instead, which, in fact, it is the well-known interest and advantage of the use of kernels. Next subsection proposes an alternative at this respect.

\subsection{A kernel definition using quadratic kernels instead of expanding the mapping function $\phi$} \label{sec:kernelq}
This section proposes an alternative expression for the kernel of that of the Equation (\ref{eq:15}), which will avoid expanding the mapping function $\phi$. Let us notice that the image space of $\phi$ is the set of monomials of degree between $0$ and $2$ formed with the components of $x$ and $s$, but removing the monomials with more than one component of $x$ or with more than one component of $s$, including the squared monomials (see the Equation (\ref{eq:11})). This fact sheds light on defining the kernel in terms of the existing quadratic kernel, which adopts the form 

\begin{equation}\label{eq:16}
K_{Q,c}(u,v)=(\langle u, v \rangle+c)^2=\left (\sum_{i=1}^n u_iv_i+c\right)^2
\end{equation}

The quadratic kernel expressed in the Equation (\ref{eq:16}) computes the inner product in the original space, hence, it is of linear order with regard to the features. The idea is to express the Equation (\ref{eq:15}) in terms of a sum or a difference of several quadratic kernels, then, obtaining an expression of linear order with regard to the features. Particularly, the alternative kernel expression proposed is the following one
\begin{multline}\label{eq:17}
K\left(\left(x^{(1)},s^{(1)}\right),\left(x^{(2)},s^{(2)}\right)\right)=\frac{1}{2}\cdot \left (K_{Q, 1}\left(\left(x^{(1)},s^{(1)}\right),\left(x^{(2)},s^{(2)}\right) \right)\right. -\\
-\left. K_{Q, 0}\left(x^{(1)},x^{(2)}\right)-K_{Q, 0}\left(s^{(1)},s^{(2)}\right)+1\right )
\end{multline}

Effectively, let us now demonstrate that the expression of the Equation (\ref{eq:17}) is equivalent to the expression of the Equation (\ref{eq:15}). For this purpose, let us first use the binomial theorem of elementary algebra to the Equation (\ref{eq:16}). Then, 
the expression of the Equation (\ref{eq:16}) becomes
\begin{equation}\label{eq:18}
K_{Q,c}(u,v)=\left(\sum_{i=1}^n u_iv_i\right)^2+2c\left(\sum_{i=1}^n u_iv_i\right)+c^2
\end{equation}
Now, let us apply the multinomial theorem to the first term of the expression of the Equation (\ref{eq:18}) and regrouping the terms properly. The result is the expression of the Equation (\ref{eq:19})
\begin{multline}\label{eq:19}
K_{Q,c}(u,v)=\sum_{i=1}^n (u_i^2)(v_i^2) + \sum_{i=2}^n\sum_{j=1}^{i-1}(\sqrt{2}u_iu_j)(\sqrt{2}v_iv_j) + \\
+\sum_{i=1}^n(\sqrt{2c}u_i)(\sqrt{2c}v_i)+c^2
\end{multline}

Next, and using the expression of the Equation (\ref{eq:19}), let us expand the terms of the Equation (\ref{eq:17}) 
\begin{itemize}
\item [i)]$K_{Q, 1}\left(\left(x^{(1)},s^{(1)}\right),\left(x^{(2)},s^{(2)}\right)\right)$, that is, for $c=1$, $u=\left(x^{(1)},s^{(1)}\right)$ and $v=\left(x^{(2)},s^{(2)}\right)$
\item [ii)]$K_{Q, 0}\left(x^{(1)}\right.,$ $\left. x^{(2)}\right)$, that is, for $c=0$, $u=x^{(1)}$ and $v=x^{(2)}$
\item[iii)]$K_{Q, 0}\left(s^{(1)},s^{(2)}\right)$, that is, for $c=0$, $u=s^{(1)}$ and $v=s^{(2)}$
\end{itemize}

 Then, expanding these terms will respectively lead to the Equations (\ref{eq:20}), (\ref{eq:21}) and (\ref{eq:22}).
\begin{multline}\label{eq:20}
K_{Q, 1}\left(\left(x^{(1)},s^{(1)}\right),\left(x^{(2)},s^{(2)}\right)\right)=\left (\langle  (x^{(1)},s^{(1)}), (x^{(2)},s^{(2)})\rangle+1\right )^2=\\
 =\sum_{i=1}^{a_x}\left(x_i^{(1)}\right)^2 \left(x_i^{(2)}\right)^2+\sum_{i=1}^{a_s}\left(s_i^{(1)}\right)^2 \left(s_i^{(2)}\right)^2+\\
+\sum_{i=2}^{a_x}\sum_{j=1}^{i-1}\left(\sqrt{2} x_i^{(1)}x_j^{(1)}\right)\left(\sqrt{2} x_i^{(2)}x_j^{(2)}\right)+\\
+\sum_{i=2}^{a_s}\sum_{j=1}^{i-1}\left(\sqrt{2} s_i^{(1)}s_j^{(1)}\right)\left(\sqrt{2} s_i^{(2)}s_j^{(2)}\right)+\\
+\sum_{i=1}^{a_x}\sum_{j=1}^{a_s}\left(\sqrt{2} x_i^{(1)}s_j^{(1)}\right)\left(\sqrt{2} x_i^{(2)}s_j^{(2)}\right)+\\
+\sum_{i=1}^{a_x}\left(\sqrt{2}  x_i^{(1)}\right) \left (\sqrt{2}x_i^{(2)}\right)+\sum_{i=1}^{a_s}\left(\sqrt{2}  s_i^{(1)}\right) \left (\sqrt{2}s_i^{(2)}\right)+1
\end{multline}
\begin{multline}\label{eq:21}
K_{Q, 0}\left(x^{(1)},x^{(2)}\right)=\left(\langle x^{(1)}, x^{(2)}\rangle+0\right)^2=\sum_{i=1}^{a_x}\left(x_i^{(1)}\right)^2 \left(x_i^{(2)}\right)^2+\\
 +\sum_{i=2}^{a_x}\sum_{j=1}^{i-1}\left(\sqrt{2} x_i^{(1)}x_j^{(1)}\right)\left(\sqrt{2} x_i^{(2)}x_j^{(2)}\right)
\end{multline}

\begin{multline}\label{eq:22}
K_{Q, 0}\left(s^{(1)},s^{(2)}\right)=\left(\langle s^{(1)}, s^{(2)}\rangle+0\right)^2=\sum_{i=1}^{a_s}\left(s_i^{(1)}\right)^2 \left(s_i^{(2)}\right)^2+\\
 +\sum_{i=2}^{a_s}\sum_{j=1}^{i-1}\left(\sqrt{2} s_i^{(1)}s_j^{(1)}\right)\left(\sqrt{2} s_i^{(2)}s_j^{(2)}\right)
\end{multline}

Therefore, replacing the terms of  $K_{Q, 1}\left(\left(x^{(1)},s^{(1)}\right),\left(x^{(2)},s^{(2)}\right)\right)$, $K_{Q, 0}\left(x^{(1)},\right.$ $\left. x^{(2)}\right)$ and $K_{Q, 0}\left(s^{(1)},s^{(2)}\right)$ in the Equation (\ref{eq:17}) by the expressions of the Equations (\ref{eq:20}), (\ref{eq:21}) and (\ref{eq:22}) and  removing the terms that are annulled, $K\left(\left(x^{(1)},s^{(1)}\right),\left(x^{(2)},s^{(2)}\right)\right)$  adopts the following expression
\begin{multline}\label{eq:23}
K\left(\left(x^{(1)},s^{(1)}\right),\left(x^{(2)},s^{(2)}\right)\right)= \frac{1}{2}\cdot \left (\sum_{i=1}^{a_x}\sum_{j=1}^{a_s}\left(\sqrt{2} x_i^{(1)}s_j^{(1)}\right)\left(\sqrt{2} x_i^{(2)}s_j^{(2)}\right)+\right .\\
+ \left. \sum_{i=1}^{a_x}\left(\sqrt{2}  x_i^{(1)}\right) \left (\sqrt{2}x_i^{(2)}\right)+\sum_{i=1}^{a_s}\left(\sqrt{2}  s_i^{(1)}\right) \left (\sqrt{2}s_i^{(2)}\right)+2\right)=\\
\end{multline}
Now, simplifying the constant, the expression becomes
\begin{multline}\label{eq:24}
K\left(\left(x^{(1)},s^{(1)}\right),\left(x^{(2)},s^{(2)}\right)\right)= \sum_{i=1}^{a_x}\sum_{j=1}^{a_s}\left(x_i^{(1)}s_j^{(1)}\right)\left(x_i^{(2)}s_j^{(2)}\right)+\\
+\sum_{i=1}^{a_x}\left( x_i^{(1)}\right) \left (x_i^{(2)}\right)+\sum_{i=1}^{a_s}\left(s_i^{(1)}\right) \left (s_i^{(2)}\right)+1
\end{multline}

Finally, one can easily observe that the expressions (\ref{eq:15}) and (\ref{eq:24}) of $K\left(\left(x^{(1)},s^{(1)}\right), \left(x^{(2)},s^{(2)}\right)\right)$ are identical after adequately ordering their terms. Therefore, we have got defining a kernel using the expression of the Equation (\ref{eq:16}) through a set of quadratic kernels whose computation is of linear order with regard to the features instead of using either the expression of the Equation (\ref{eq:15}) or (\ref{eq:24}) which are of quadratic order with regard to the features.

\subsection{Toy example}\label{sec:toyexample}
This section presents a toy example that illustrates the way the method \dsil\ works. 
\begin{figure}
	\centering
	\includegraphics[width=\textwidth, trim={1.5cm 11cm 7cm 2cm}, clip]{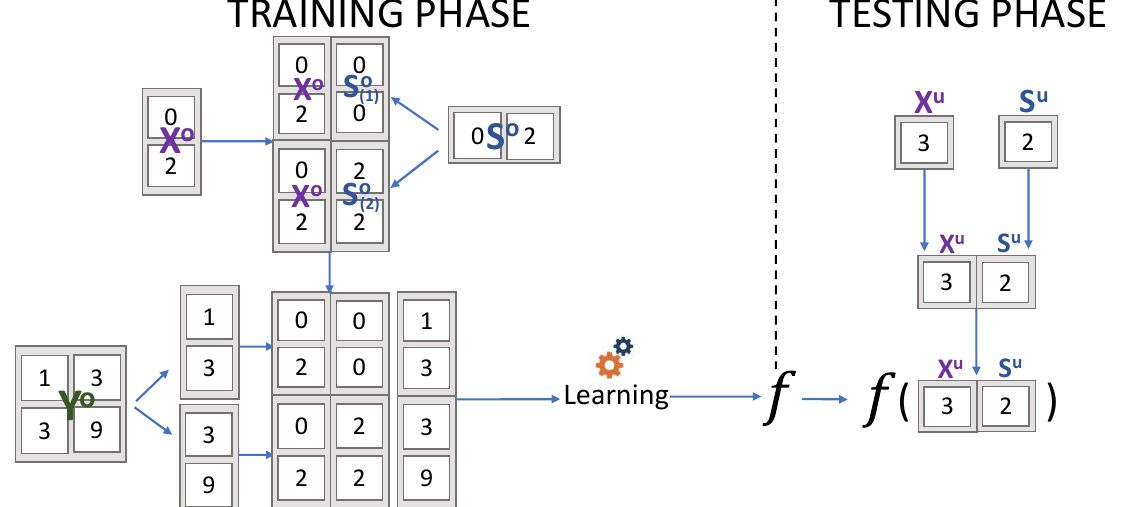}
	\caption{A toy example that illustrates the direct side information learning method}
	\label{fig:exampleExperimentDSIL}
	\centering
\end{figure}
Figure \ref{fig:exampleExperimentDSIL} shows how \dsil\ works through a toy example. Let the number of observed targets $m_o$ be equal to $2$, whose side information size $a_s$ is equal to $1$ (each target is represented by a one-dimensional vector).  Let one of the observed targets be $t^o_1=0$ and the other observed target be $t^o_2=2$ (both represented in $S^o$ in Figure \ref{fig:exampleExperimentDSIL}). Let us consider that each of these two targets has two instances (the same for simplicity), whose number of features $a_x$ is equal to $1$ (each instance is represented by a one-dimension vector). Let these two instances be $x^o_1=0$ and $x^o_2=2$ (both represented in $X^o$ in Figure \ref{fig:exampleExperimentDSIL}). Finally, let us suppose a linear relationship between features $x$ and predictions $y$, whose coefficients are in turn a linear function of the side information, that is, the linear function to be learned is $y=f(x,s)=\alpha(s)\cdot x+\beta(s)$, where $\alpha(s)$ and $\beta(s)$ are in turn linear functions. Let $\alpha(s)=s+1$ and $\beta(s)=s+1$ for simplicity. Then, the predictions will be $1$ and $3$ for the two instances of the observed target $t^o_1$ and $3$ and $9$ for the two instances of the observed target $t^o_2$ (represented in $Y^o$ in Figure \ref{fig:exampleExperimentDSIL}). Then, the learning process to induce the function model $f$ takes place. Notice that this schema also fits the method \base\, changing the kernel in the learning process. Once the function $f$ is learned, for an unobserved instance $x^u=3$ (represented by $X^u$ in Figure \ref{fig:exampleExperimentDSIL}) and for an unobserved target $t^u=2$ (represented by $S^u$ in Figure \ref{fig:exampleExperimentDSIL}) the prediction using $f$ is carried out. For this purpose, the feature description $x^u=3$ and the side information $t^u=2$ for an unobserved target is concatenated for feeding $f$ and obtaining $f(x^u, s^u)=f(3,2)$. 

\begin{figure}
	\centering
	\includegraphics[width=\textwidth, trim={0cm 6cm 0.2cm 1.6cm}, clip]{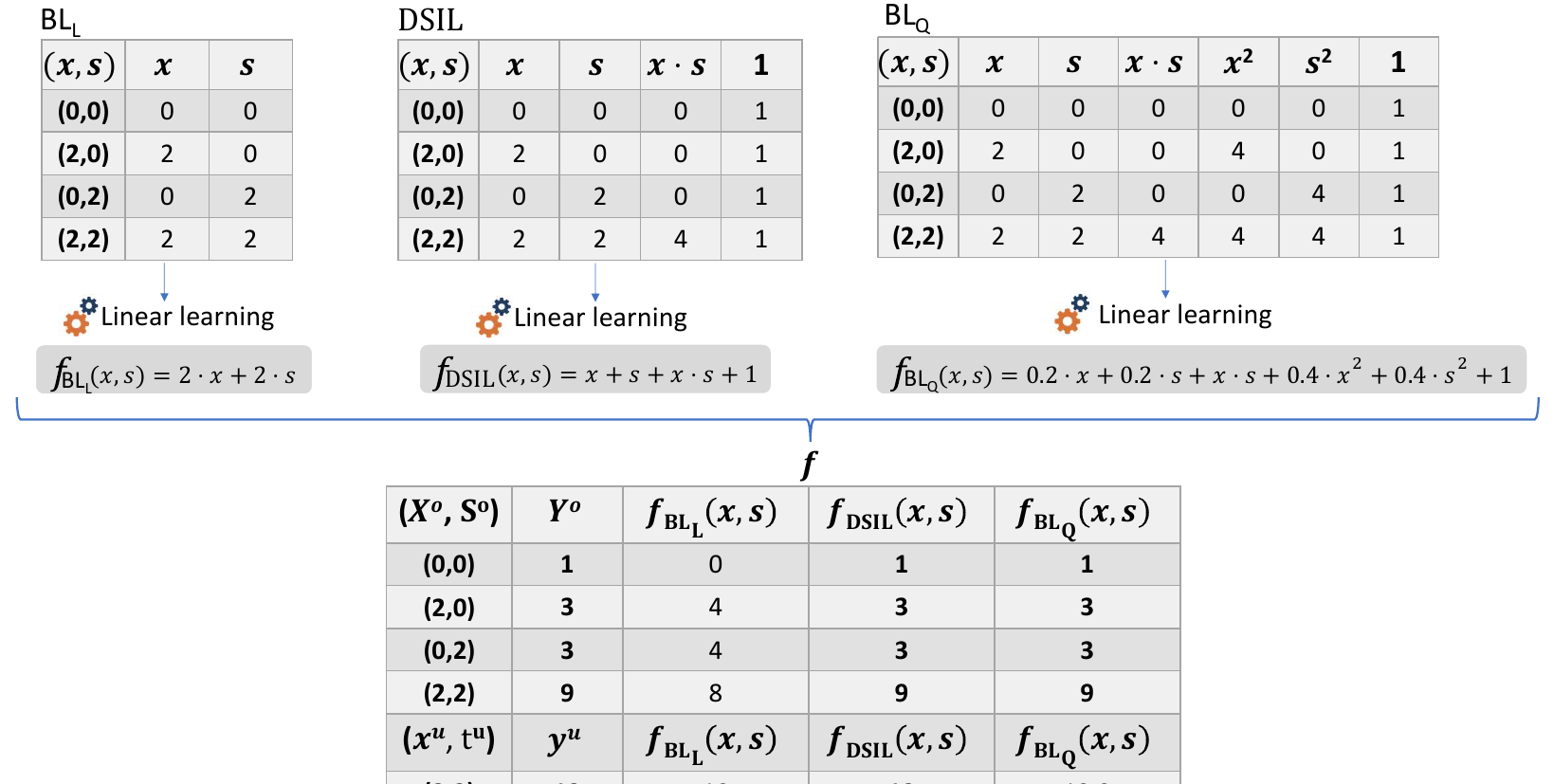}
	\caption{Differences of the methods \basel{}, \dsil{} y \basep{} through the toy example}
	\label{fig:EvaluacionDSIL}
	\centering
\end{figure}

Once the toy example is defined, let us compare the method \base\ with linear kernel (\basel), the method \dsil\ and the method \base\ with quadratic kernel (\basep). Figure \ref{fig:EvaluacionDSIL} 
displays the learning process of the three methods. Firstly, the function mapping $\phi$ is applied according to the different kernels that the three methods use. Let us remind that $a_x=a_s=1$ in the toy example. Then, both $x$ and $s$ are one-dimensioned. The mapping function $\phi(x,s)$ for the method \basel\ is $\phi(x,s)=(x,s)$, then, only the monomials $x$ and $s$ are taken. In the case of the method \dsil, the mapping function $\phi(x,s)$ equals $(x, x\cdot s, s, 1)$, then, the monomials that this kernel considers are also $x$ and $s$, but in addition, the monomials $x\cdot s$ and $1$. Finally, the mapping function $\phi(x,s)$ for the kernel of the method \basep\ is $(x^2, x, x\cdot s, s, s^2, 1)$, that is, all possible combinations of monomials of degree $2$, namely, $x^2$, $s^2$ in addition to $x$, $s$, $x\cdot s$ and $1$. Once the expansion is carried out, a linear learning takes place in the image space of the mapping function $\phi$. The learned models were respectively $f_{\text{\basel}}(x,s)=2\cdot x+2\cdot s$, $f_{\text{\dsil}}(x,s)=x+s+x\cdot s+1$ and $f_{\text{\basep}}(x,s)=0.2\cdot x+0.2\cdot s+x\cdot s+0.4\cdot x^2+0.4\cdot s^2+1$. The evaluations of these models over both the observed instances and the unobserved instance are shown on the bottom of Figure \ref{fig:EvaluacionDSIL}. This toy example shows that the model induced by \basel\ cannot predict neither the observed instances nor the unobserved instance properly. It also shows that the model induced by \basep\ could not have enough generalization power, since it correctly predicts the observed instances of the observed targets, but it fails in the prediction of the unobserved instance of the unobserved target (it overfits).

\subsection{Computational complexity analysis}\label{sec:complexity}
Three different implementations of \dsil\ have been proposed, namely, directly using the mapping function $\phi$ (\dsilm), defining a kernel using the mapping function $\phi$ (\dsilk) and defining a kernel that uses quadratic kernels instead of expanding the mapping function $\phi$ (\dsilkq). All three have been shown to be equivalent, since they induce the same models for the function $f$. Hence, the performance in terms of accuracy is equal, but they highly differ in terms of computational cost. This section analyses this issue. 

\dsilk\ and \dsilkq\ need the same storage requirements than \base. However, \dsilm\ requires more storage than \base\ due to the increasing number of features. Particularly, the number of features changes from $a_x+a_s$ (linear) in the case of \base\ to $(a_x+1)\cdot (a_s+1)$ (quadratic) in the case of \dsilm\ due to the mapping function $\phi$ computation (let us remind that $a_x$ and $a_s$ respectively are the instance feature and side information sizes).

Regarding time complexity, the key of the differences is focused on the kernel evaluation. The kernel is evaluated per each pair of instances. Hence, the time complexity will be $\mathcal{O}(n_o^2)\cdot \mathcal{O}(K)$, where $n_o$ is the number of instances and $K$ is the kernel. Then, let us now analyze the different methods according to the kernel they use:

\begin{itemize}
\item [i)]The \base\ method when a kernel of linear time complexity is taken, like linear kernel (\basel) or quadratic kernel (\basep), has the following time complexity  
$$\mathcal{O}(\textrm{\base})=\mathcal{O}(n_o^2)\cdot \mathcal{O}(a_x+a_s)=\mathcal{O}(n_o^2\cdot (a_x+a_s))$$ 
\item [ii)]\dsilm\ has in addition a preprocess apart from the kernel evaluations. This preprocess consists of expanding each instance using the mapping function $\phi$ obtaining an instance in the image space of $\phi$ of size $(a_x+1)\cdot(a_s+1)$. Then, the kernel evaluations will take place in the image space of $\phi$. Hence, the time complexity will have two terms, one for the preprocess $\mathcal{O}(n_o\cdot ((a_x+1)\cdot(a_s+1)))$ and the other for the kernel evaluations $\mathcal{O}(n_o^2)\cdot \mathcal{O}((a_x+1)\cdot(a_s+1))$. Therefore, the time complexity of \dsilm\ will be 
\begin{multline*}
\mathcal{O}(\textrm{DSIL}_{\phi})=\mathcal{O}(n_o\cdot ((a_x+1)\cdot(a_s+1)))+\mathcal{O}(n_o^2)\cdot \mathcal{O}((a_x+1)\cdot(a_s+1))=\\
=\mathcal{O}(n_o\cdot a_x\cdot a_s)+\mathcal{O}(n_o^2\cdot a_x\cdot a_s)=\mathcal{O}(n_o^2\cdot a_x \cdot a_s)
\end{multline*}
\item [iii)] \dsilk\ expands each instance in the kernel evaluation. Hence, the time complexity of the kernel evaluation is $\mathcal{O}((a_x+1)\cdot(a_s+1))$. Then the time complexity of \dsilk\ will be
$$\mathcal{O}(\textrm{DSIL}_{K_{\phi}})=\mathcal{O}(n_o^2)\cdot \mathcal{O}((a_x+1)\cdot(a_s+1))=\mathcal{O}(n_o^2\cdot a_x\cdot a_s)$$
\item [iv)] \dsilkq\ applies the quadratic kernel, which is of the linear order with regard to the features. It evaluates this kernel three times, respectively taken $a_x+a_s$, $a_x$ and $a_s$ features. Then, the time complexity of the kernel evaluation will have three terms $\mathcal{O}(a_x+a_s)$, $\mathcal{O}(a_x)$ and $\mathcal{O}(a_s)$. Therefore, the time complexity of \dsilkq\ will be
$$\mathcal{O}(\textrm{DSIL}_{K_{Q}})=\mathcal{O}(n_o^2)\cdot (\mathcal{O}(a_x+a_s)+\mathcal{O}(a_x) +\mathcal{O}(a_s))=\mathcal{O}(n_o^2\cdot(a_x+a_s))$$
\end{itemize}

\begin{table}[h]
	\begin{center}
	\begin{tabular}{ lllll} 
			\toprule
			\multirow{2}{*}{Algorithm} & \multirow{2}{*}{Preprocess} & \centering Kernel  & \centering Kernel& \multirow{2}{*}{Time complexity}\\
			& &\centering size& \centering evaluation &\\
			\midrule
			\basep & - &$ \mathcal{O}(n_o^2)$ &$\mathcal{O}(a_x+a_s)$&$\mathcal{O}(n_o^2\cdot(a_x+a_s))$\\
			\dsilm\ &$\mathcal{O}(n_o\cdot (a_x+a_s))$ &$ \mathcal{O}(n_o^2)$ &$\mathcal{O}(a_x\cdot a_s)$&$\mathcal{O}(n_o^2\cdot a_x\cdot a_s)$\\
			\dsilk\ & - &$ \mathcal{O}(n_o^2)$ &$\mathcal{O}(a_x\cdot a_s)$&$\mathcal{O}(n_o^2\cdot a_x \cdot a_s)$\\
			\dsilkq\ & - &$ \mathcal{O}(n_o^2)$ &$\mathcal{O}(a_x+a_s)$&$\mathcal{O}(n_o^2\cdot(a_x+a_s))$\\
			\bottomrule
		\end{tabular}
		\caption{Summary of time complexity of \basep, \dsilm, \dsilk\ and \dsilkq}
		\label{tab:complejidad}
	\end{center}
\end{table}

Table \ref{tab:complejidad} displays a summary of the time complexity of the algorithms. On the one hand, the use of the mapping function $\phi$ in the kernel evaluation (\dsilm\ and \dsilk) involves increasing the time complexity order from linear to quadratic with regard to the number of features $a_x$ and $a_s$. The expansion that takes place in the preprocess of \dsilm\ is of linear order with regard to the number of instances. \dsilk\ avoids this preprocess of \dsilm, but this affects the kernel evaluation of \dsilk\ in that \dsilk\ must perform the expansion per kernel evaluation, which is of quadratic order with regard to the number of instances. Hence, embedding the mapping function $\phi$ in the kernel evaluation does not help to improve the time complexity; otherwise, it worsens it, despite both \dsilm\ and \dsilk\ have the same time complexity order. On the other hand, \dsilkq\ is of the same order that \basep. Hence, it was possible to build a specific kernel for properly treating the side information without increasing the time complexity order. Later on, the experiments will show the improved performance that \dsilkq\ exhibits with regard to \basep.


\section{Experiments} \label{sec:experiments}
This section describes the experiments that were carried out in order to compare the approaches. Particularly, the \base{}, \sr{}, \mplc{} and \dsil{} approaches were compared. Besides, both implementations of \dsil\ were also compared in terms of computational time. Section \ref{datasets} detailed the datasets taken for the experiments. In Section \ref{settings} the parameter settings are established. Finally, Section \ref{results} analyses and discusses the results. 

\subsection{Description of datasets}\label{datasets}
There hardly are available datasets with side information in the literature for zero-shot regression. Despite several applications can fit this scenario, as it was illustrated in Section \ref{Introduction}, the problem is that the benchmark datasets do not include side information because it could be ignored or even it has not been collected.  Hence, and before applying the methods to the air pollution datasets from the Principality of Asturias of Spain, some artificial datasets were designed in order to exhaustively analyze the performance of the methods. The number of targets and the side information size were varied to build different artificial datasets in order to study the effect of the methods under different values of these parameters.

\subsubsection{Artificial datasets}
The artificial datasets are built following the same guidelines of \cite{fdez2022target}. They only differ in the number of instances in order to make the \basep{} computationally feasible, but the conclusions reported in \cite{fdez2022target} remains. A uniform distribution in the range of $(-2;-1] \cup [+1;+2)$ was taken to provide the feature values of the instances $x$ and the side information $s$ (feature values of the target description)  due to avoid zero values or values near zero. Then, given the features of an instance $x=(x_i,\dots,x_{a_x})$ and the side information of a target $s=(s_i,\dots,s_{a_s})$, prediction value $y$ for this instance is built as a linear function of $x$, whose coefficients $\{\alpha_i(s)\}_{i=1}^{a_x}$ are in turn functions of $s$, that is,
$$y=\sum_{i=1}^{a_x}(\alpha_i(s)\cdot x_i)+\beta$$

Two different ways of obtaining $\{\alpha_i(s)\}_{i=1}^{a_x}$ coefficients have been addressed in order to cover the domains that the methods \sr, \mplc{} and \dsil{} are able to encompass\footnote{This process has not been done for the \base{} method, since it would mean that the side information would not reproduce the relationship between the instances and targets; otherwise it would lead to a general-purpose regression task and not a general-purpose zero-shot regression task.}. On the one hand, datasets will simulate the most general structure that provides a linear dependence of side information. On the other hand, the target will be generated in terms of a similarity measure $\delta$ applied to the side information $s$ of the target and a set of other side information descriptions $\{\mu^k\}_{k=1}^d$(see \cite{fdez2022target} for more details).
Formally, 

$$\alpha_i(s)=\sum_{j=1}^{a_s}(\gamma_{i,j}\cdot s_j)+\beta_i \qquad \alpha_i(s)=\frac{\sum_{k=1}^{d} \left(\tau_{i,k}\cdot \delta(s,\mu^k) \right)}{\sum_{k=1}^{d}\delta(s,\mu^k)} $$

The same uniform distribution in the range of $(-2;-1] \cup [+1;+2)$ as for the feature values of $X$ and $S$ was taken for obtaining $\beta$, $\{\beta_i\}_{i=1}^{a_s}$, $\{\gamma_{i,j}\}_{i=1, j=1}^{a_x, a_s}$, $\{\tau_{i,k}\}_{i=1, k=1}^{a_x, d}$ and $\{\mu^{k}\}_{k=1}^{d}$ coefficients. The similarity function $\delta$ has been randomly chosen to be either the Manhattan (L1 norm) or the Euclidean (L2 norm) in equal shares in order to avoid bias.

The number of instances $n_o$ and features $a_x$ were respectively fixed to $500$ ($5000$ in \cite{fdez2022target}) and $50$ (the same value as in \cite{fdez2022target}) in both ways of obtaining $\{\alpha_i(s)\}_{i=1}^{a_x}$. The number of targets $m_o$ considered were ${5, 10, 50}$ and ${100}$, and the side information size $a_s$ were ${5, 15}$ and ${25}$, in order to cover a range both below and above the real datasets (the same values as in \cite{fdez2022target}). Table \ref{tab:artificialdatasets} shows the artificial datasets taken. The $\text{R}^{k,l}$  datasets reproduce a linear dependence of the side information, whereas $\text{S}^{k,l}$ datasets simulate a similarity measure, where $k$ and $l$ respectively are the number of targets $m_o$ and the side information size $a_s$.

\begin{table}[H]
	\begin{center}
	\begin{tabular}{ lrc | lrc  | lrc} 
			\toprule
			dataset & $m_o$ & $a_s$ &dataset & $m_o$ & $a_s$ & dataset & $m_o$ & $a_s$\\
			\midrule
			$\{\text{R, S}\}^{5,5}$ & 5 & 5 &$\{\text{R, S}\}^{5,15}$ &  5 & 15 &$\{\text{R, S}\}^{5,25}$ & 5 & 25  \\
			$\{\text{R, S}\}^{10,5}$ &  10 & 5 &$\{\text{R, S}\}^{10,15}$ & 10 & 15 &$\{\text{R, S}\}^{10,25}$ & 10 & 25 \\
			$\{\text{R, S}\}^{50,5}$ & 50 & 5 &$\{\text{R, S}\}^{50,15}$ & 50 & 15 & $\{\text{R, S}\}^{50,25}$ &  50 & 25 \\
			$\{\text{R, S}\}^{100,5}$ & 100 & 5 &$\{\text{R, S}\}^{100,15}$ &  100 & 15 &$\{\text{R, S}\}^{100,25}$ & 100 & 25 \\
			\bottomrule
		\end{tabular}
		\caption{Number of targets and side information size of the artificial datasets. The $\text{R}^{k,l}$  datasets reproduce a linear dependence of the side information, whereas $\text{S}^{k,l}$ datasets simulate a similarity measure, where $k$ and $l$ respectively are the number of targets and the side information size.}
		\label{tab:artificialdatasets}
	\end{center}
\end{table}

Other artificial datasets were generated in order to compare the computational time of the different implementations of \dsil. Let us make vary the number of instances ($n_o$) and features ($a_x$) together with the number of targets ($m_o$) and side information size ($a_s$). Let us remind that the number of features that feed the \dsil\ approach is the number of features ($a_x$) plus the side information size ($a_s$), that is, $a_x+a_s$. Besides, the number of instances that feed the \dsil\ approach is the number of instances ($n_o$) multiplied by the number of targets ($m_o$), that is $n_o\cdot m_o$ (assuming that the same instances are shared by all the targets, for simplicity). Hence, a range of representative values for each $a_x$, $a_s$, $n_o$ and $m_o$ were taken to perform the comparison. Particularly, both $a_x$ and $a_s$ took values $10$, $100$, $250$ and $500$, whereas the values for $n_o$ were $10$, $20$, $30$ and $40$ and the values for $m_o$ were $5$, $10$, $15$ and $20$. Then, the values for the number of features that feed the \dsil\ approach ($a_x+a_s$) were $20$, $200$, $500$ and $1000$, whereas the values for the number of instances ($n_o\cdot m_o$) were $50$, $200$, $450$ and $800$. 

\subsubsection{Air pollution datasets}
\begin{table}[H]
	\begin{center}
		\begin{tabular}{ l | r | r | r | c } 
		\toprule
                         \multirow{1}{*}{dataset}&\multirow{1}{*}{$n_o$}&\multirow{1}{*}{$a_x$}&\multirow{1}{*}{$m_o$}&\multirow{1}{*}{$a_s$}\\
  			\midrule
			NO$_2$, PST, NO, SO$_2$, CO, O$_3$ &41325& 12& 5 & 16 \\			
		\bottomrule
		\end{tabular}
		\caption{Properties of the air pollution datasets}
		\label{tab:realdatasets}
	\end{center}
\end{table}

A total of 11 pollutants are collected every 15 minutes (and hourly averaged) between 2010 and 2018 in 18 pollution and meteorological stations located in the Principality of Asturias, Spain. However, just 6 pollutants (NO$_2$, PST, NO, SO$_2$, CO and O$_3$) and 5 stations ($m_o=5$) are taken to get the largest common set of pollutants and stations, since not all stations collect all the pollutants. The instance features are a total of $12$ ($a_x=12$) hourly averaged weather conditions, such as wind direction according to Cartesian axes, the season or the precipitation. The goal is to predict the hourly pollutant concentration. The stations are the targets for which surrounding characteristics were collected to form the side information. The fact is that environmental experts \cite{delavar2019novel, murillo2019forecasting} argue that pollutant concentration not only depends on the weather conditions; otherwise it is highly influenced by the environment where it is gathered. Particularly, the surrounding information collected consisted of establishing if an urban center or highway or factory or sea (or river) is nearby the station in the North, South, East and/or West directions. Then, a total of 16 features (4 cardinal points multiplied by 4 possible kinds of surroundings) have formed the side information ($a_s=16$). Hence, the goal is to predict pollutant concentration from certain weather conditions taken in a station for which weather conditions have not been collected (unobserved station). Table \ref{tab:realdatasets} displays the properties of the $6$ pollutants  (NO$_2$, PST, NO, SO$_2$, CO and O$_3$) datasets.
			
\subsection{Parameter settings} \label{settings}
All the approaches were implemented in Python using \textit{Scikit Learn Library} \footnote{\url{https://scikit-learn.org/stable/}}.  All the code is available in \href{https://github.com/UO231492/UO231492/DSILZSR}{\textit{GitHub repository}}\footnote{\url{https://github.com/UO231492/DSILZSR}}. Support Vector Regression (SVR) \footnote{\url{https://scikit-learn.org/stable/modules/generated/sklearn.svm.SVR.html}} is required in order to make possible the use of a quadratic kernel in the \base\ approach and also to implement the versions of \dsil\ that involve kernels. Hence, SVR is taken for all the approaches in order to get a fair comparison. \basel\ and \basep\ are the \base\ approach respectively using linear and quadratic kernels. Only these kernels are considered, since DSIL is in between both, that is, DSIL considers the monomials of order $1$, as the linear kernel, but it also includes some monomials of order $2$, but not all possible monomials of order $2$ that the quadratic kernel includes. \sre\ and \srm\ are the \sr\ approach respectively using the typical Euclidean (L2 norm) and Manhattan (L1 norm) distances to define the relationship between observed and unobserved targets. \mplc\ has just one version. Finally, \dsil\ admits three different implementations, namely \dsilm, \dsilk\ and \dsilkq, which respectively are the implementation that computes $\phi$ mapping directly, the implementation that computes the kernel using $\phi$ mapping and the implementation that computes the kernel though a linear combination of quadratic kernels.  However, the three implementations report the same performance scores since, in fact, it is the same method. Then, they just differ in the computational time that will be also compared. The hyperparameter $c$ of SVR was optimized through a grid search procedure taking the values $\{10^{-3},10^{-2},\dots, 10^{2}, 10^{3}\}$ for $c$ and using a 3-fold cross validation for estimating the mean squared error. The score for comparing the approaches was the relative mean squared error computed following again a 3-fold cross validation. The cross validation performed under the existence of side information is slightly different from the common cross validation. Under this paradigm, not only instances are taken into account; otherwise, both instances and targets must be considered. Besides, some prediction values for the testing fold must be discarded for consistency. These values are i) for the observed (training) targets and unobserved (testing) instances and ii) for the unobserved (testing) targets and observed (training) instances (see \cite{fdez2022target} for more information). None cross-validation was performed for comparing the computational time of the \dsil\ approaches (\dsilm, \dsilk\ and \dsilkq) in order to avoid the little differences in number of instances and targets that may be in the different folds. A training-test procedure repeated 3 times was carried out instead. A Friedman-Nemeyi test \cite{demsar2006statistical} was performed. The Friedman test is a non-parametric hypothesis test that ranks all algorithms for each data set separately. If the null-hypothesis (all ranks are not significantly different) is rejected, the Nemenyi test is adopted as the post-hoc test. According to the Nemenyi test, the performance of two algorithms is considered significantly different if the corresponding average ranks differ by at least the so-called critical difference. 
\begin{table}[!p]
	\resizebox{12cm}{!}{
		\begin{tabular}{lrrr}
			\toprule
			Dataset&\basel&\basep&\dsil\\ \hline
    	    $\text{R}^{5,5}$&112.03(2)&113.85(3)&\textbf{58.55}(1)\\ 
            $\text{R}^{10,5}$&131.46(3)&0.04(2)&\textbf{3.32E-12}(1)\\ 
            $\text{R}^{50,5}$&108.35(3)&4.70E-05(2)&\textbf{8.61E-15}(1)\\ 
            $\text{R}^{100,5}$&111.40(3)&2.58E-05(2)&\textbf{2.54E-15}(1)\\ 
            $\text{R}^{5,15}$&115.68(3)&89.13(2)&\textbf{78.73}(1)\\ 
            $\text{R}^{10,15}$&184.42(3)&72.66(2)&\textbf{61.71}(1)\\ 
            $\text{R}^{50,15}$&100.68(3)&2.02E-05(2)&\textbf{2.00E-14}(1)\\ 
            $\text{R}^{100,15}$&117.02(3)&1.30E-05(2)&\textbf{3.21E-15}(1)\\ 
            $\text{R}^{5,25}$&\textbf{104.87}(1)&107.74(3)&105.50(2)\\ 
            $\text{R}^{10,25}$&194.14(3)&80.82(2)&\textbf{65.25}(1)\\ 
            $\text{R}^{50,25}$&102.47(3)&1.11E-04(2)&\textbf{6.47E-13}(1)\\ 
            $\text{R}^{100,25}$&108.13(3)&1.34E-05(2)&\textbf{4.89E-15}(1)\\ 
			\midrule
			Avg. Rank&\ \ \  \ (2.75)&\ \ \  \ (2.17)&\ \ \  \ (\textbf{1.08})\\
			\midrule
				Dataset&\basel&\basep&\dsil\\ 
				\midrule
    	$\text{S}^{5,5}$&\textbf{6.07}(1)&114.58(2)&136.56(3)\\ 
        $\text{S}^{10,5}$&2.97(2)&76.97(3)&\textbf{2.45}(1)\\ 
        $\text{S}^{50,5}$&1.74(2)&85.13(3)&\textbf{0.41}(1)\\ 
        $\text{S}^{100,5}$&1.42(2)&73.45(3)&\textbf{0.20}(1)\\ 
        $\text{S}^{5,15}$&\textbf{0.20}(1)&107.15(3)&63.40(2)\\ 
        $\text{S}^{10,15}$&\textbf{0.33}(1)&87.87(3)&70.75(2)\\ 
        $\text{S}^{50,15}$&0.31(2)&89.43(3)&\textbf{0.04}(1)\\ 
        $\text{S}^{100,15}$&0.38(2)&93.31(3)&\textbf{0.03}(1)\\ 
        $\text{S}^{5,25}$&\textbf{0.63}(1)&69.63(3)&54.71(2)\\ 
        $\text{S}^{10,25}$&\textbf{1.96}(1)&71.69(3)&54.55(2)\\ 
        $\text{S}^{50,25}$&1.70(2)&83.00(3)&\textbf{0.50}(1)\\ 
        $\text{S}^{100,25}$&1.21(2)&89.55(3)&\textbf{0.16}(1)\\ 
			\midrule
			Avg. Rank&\ \ \  \ (1.58)&\ \ \  \ (2.92)&\ \ \  \ (\textbf{1.50})\\
			\midrule\midrule
			Mean Rank&\ \ \  \ (2.17)&\ \ \  \ (2.55)&\ \ \  \ (\textbf{1.29})\\
			\bottomrule
		\end{tabular}
	}
	\caption{Mean relative square error and Friedman ranks for artificial datasets using \basel,  \basep{} and \dsil{}.}
	\label{tab:artificialBLDSIL}
\end{table}

\subsection{Result analysis and discussion} \label{results}
This section analyses and discusses the results of the experiments. 

\subsubsection{Artificial datasets}
\begin{table}[!p]
	\resizebox{12cm}{!}{
		\begin{tabular}{lrrrr}
			\toprule
			Dataset&\sre&\srm&\mplc&\dsil\\ \hline
    		$\text{R}^{5,5}$&105.76(3)&102.90(2)&111.74(4)&\textbf{58.55}(1)\\ 
            $\text{R}^{10,5}$&115.20(4)&105.75(3)&3.61E-08(2)&\textbf{3.32E-12}(1)\\ 
            $\text{R}^{50,5}$&72.04(4)&67.05(3)&1.84E-10(2)&\textbf{8.61E-15}(1)\\ 
            $\text{R}^{100,5}$&78.90(4)&72.72(3)&9.79E-11(2)&\textbf{2.54E-15}(1)\\ 
            $\text{R}^{5,15}$&105.26(4)&104.94(3)&88.97(2)&\textbf{78.73}(1)\\ 
            $\text{R}^{10,15}$&197.03(3)&197.18(4)&85.73(2)&\textbf{61.71}(1)\\ 
            $\text{R}^{50,15}$&93.10(4)&89.17(3)&2.97E-10(2)&\textbf{2.00E-14}(1)\\ 
            $\text{R}^{100,15}$&104.29(4)&101.45(3)&1.06E-10(2)&\textbf{3.21E-15}(1)\\ 
            $\text{R}^{5,25}$&\textbf{101.56}(1)&101.60(2)&107.88(4)&105.50(3)\\ 
            $\text{R}^{10,25}$&157.36(4)&156.97(3)&87.09(2)&\textbf{65.25(1)}\\ 
            $\text{R}^{50,25}$&100.03(4)&97.81(3)&6.03E-09(2)&\textbf{6.47E-13}(1)\\ 
            $\text{R}^{100,25}$&102.22(4)&100.21(3)&1.22E-10(2)&\textbf{4.89E-15}(1)\\ 
			\midrule
			Avg. Rank&\ \ \  \ (3.58)&\ \ \  \ (2.92)&\ \ \  \ (2.33)&\ \ \  \ (\textbf{1.17})\\
						\midrule
			Dataset&\sre&\srm&\mplc&\dsil\\ 
						\midrule
$\text{S}^{5,5}$&\textbf{6.73}(1)&6.84(2)&136.56(3.5)&136.56(3.5)\\ 
$\text{S}^{10,5}$&\textbf{2.30}(1)&2.33(2)&6.79(4)&2.45(3)\\ 
$\text{S}^{50,5}$&1.33(4)&1.30(3)&\textbf{0.28}(1)&0.41(2)\\ 
$\text{S}^{100,5}$&1.10(4)&1.08(3)&\textbf{0.20}(1.5)&\textbf{0.20}(1.5)\\ 
$\text{S}^{5,15}$&\textbf{0.19}(1)&0.20(2)&109.21(4)&63.40(3)\\ 
$\text{S}^{10,15}$&\textbf{0.30}(1.5)&\textbf{0.30}(1.5)&70.73(3)&70.75(4)\\ 
$\text{S}^{50,15}$&0.29(3.5)&0.29(3.5)&\textbf{0.04}(1.5)&\textbf{0.04}(1.5)\\ 
$\text{S}^{100,15}$&0.35(4)&0.34(3)&\textbf{0.03}(1.5)&\textbf{0.03}(1.5)\\ 
$\text{S}^{5,25}$&\textbf{0.65}(1)&0.66(2)&54.71(3.5)&54.71(3.5)\\ 
$\text{S}^{10,25}$&\textbf{1.90}(1.5)&\textbf{1.90}(1.5)&54.55(3.5)&54.55(3.5)\\ 
$\text{S}^{50,25}$&1.52(4)&1.51(3)&\textbf{0.50}(1.5)&\textbf{0.50}(1.5)\\ 
$\text{S}^{100,25}$&1.16(4)&1.15(3)&\textbf{0.16}(1.5)&\textbf{0.16}(1.5)\\ 
				\midrule
			Avg. Rank&\ \ \  \ (2.54)&\ \ \  \ (\textbf{2.46})&\ \ \  \ (2.50)&\ \ \  \ (2.50)\\			
			\midrule
			\midrule
			Mean Rank&\ \ \  \ (3.06)&\ \ \  \ (2.69)&\ \ \  \ (2.42)&\ \ \  \ (\textbf{1.84})\\ 
			\toprule
		\end{tabular}
	}
\caption{Mean relative square error and Friedman ranks for artificial datasets using \sre, \srm, \mplc, and \dsil{}.}
	\label{tab:artificialSRMPLCDSIL}
\end{table}

Table \ref{tab:artificialBLDSIL} displays the mean relative square error and Friedman ranks for artificial datasets using \basel{}, \basep{} and \dsil. The interest of this comparison lies in finding out to what extent the \dsil\ approach reproduces a linear relationship between features and targets with regard to a linear approach (\basel) and a quadratic approach (\basep).  Clearly, DSIL outperforms both \basel\ and \basep. Besides, the differences are statistically significant at the level of $95\%$ in the case of the \basep\ for both kinds of artificial datasets and in the case of \basel\ for $R^{k,l}$ datasets, since the critical difference of the Nemenyi test at this level is $0.95$. The exception is with regard to \basel\ for $S^{k,l}$ datasets, for which \dsil\ and \basel\ perform highly similar. In this case, \dsil\ and \basel\ split up the best performance. \dsil\ outperforms \basel\ when there are many targets, whereas \basel\ gets the best performance when the number of targets is low. Globally, \dsil\ is also statistically better than both \basel\ and \basep\ at the significant level of $95\%$ with a critical difference of the Nemenyi test of $0.67$. 

Table \ref{tab:artificialSRMPLCDSIL} displays the mean relative square error and Friedman ranks for artificial datasets using \sre, \srm, \mplc, and \dsil{}. \dsil\ exhibits the best performance for $R^{l,k}$ datasets. Besides, \dsil\ is significantly better than both \sre\ and \srm\ at the significant level of $95\%$, since the critical difference is $1.35$. However, all the approaches perform highly similar for $S^{l,k}$ datasets, where \srm\ is slightly better than the rest. In certain sense, this result is expected because of the way the artificial datasets were built. In any case, \dsil\ continues providing the best global performance. However, the differences in global performance are only significant in the case of \sre\ at the level of $95\%$, for which the critical difference is $0.95$. Reducing the significant level to $90\%$, \dsil\ provides significant differences with regard to both \sre\ and \srm\ because the critical difference in this case is $0.85$. Finally, there are no significant differences between \dsil\ and \mplc.	
	
\subsubsection{Real datasets}
Table \ref{tab:real} exhibits the mean relative square error and  Friedman ranks for air pollution datasets using \basel, \basep, \sre, \srm, \mplc{} and \dsil{}. Clearly, \dsil\ provides the best performance for most of the real datasets. The exceptions are the SO$_2$ and O$_3$ pollutant dataset, for which \srm\ slightly outperforms \dsil. The critical differences at the level of $99\%$, $95\%$ and $90\%$ respectively are $3.63$, $3.07$ and $2.79$. Hence, \dsil\ is significantly better than \basep\ at the level of $90\%$ and $95\%$ and even of $99\%$. Despite \dsil\ outperforms all the methods, \srm\ and \mplc\ also reach good performance with regard to \basep, \basel\ and \sre. These results confirm the hypothesis that properly treating side information will lead to an improvement in performance. However, as it can be seen, it varies from one pollutant to another, probably because the relationship between features and targets and between this relationship and the side information varies from one pollutant to another.

\begin{table}
\begin{tabular}{lllllll}
\hline
Dataset&\basel&\basep&\sre&\srm&\mplc&\dsil\\ \hline
NO$_2$&86.11(5)&117.49(6)&81.07(4)&76.52(3)&76.43(2)&\textbf{71.70}(1)\\ 
PST&90.86(4)&100.42(6)&91.17(5)&88.95(3)&87.15(2)&\textbf{86.64}(1)\\ 
NO&96.20(5)&98.71(6)&93.84(4)&90.28(2)&91.68(3)&\textbf{89.45}(1)\\ 
SO$_2$&95.01(4)&100.95(6)&95.53(5)&\textbf{92.19}(1)&93.49(3)&92.87(2)\\ 
CO&90.08(2)&100.11(6)&99.27(5)&98.48(4)&90.22(3)&\textbf{85.36}(1)\\ 
O$_3$&68.91(2)&100.50(6)&69.83(3)&\textbf{68.10}(1)&90.50(5)&76.79(4)\\ 
\hline
Avg. Rank&\ \  (3.67)&\  \ (6.00)&\ \ (4.33)& \  \ (2.33)&\  \ (3.00)&\  \ (\textbf{1.66})\\ \hline
\end{tabular}
	\caption{Mean relative square error and Friedman ranks for air pollution datasets using \basel, \basep, \sre, \srm, \mplc{} and \dsil{}.}
	\label{tab:real}
\end{table}

\begin{figure}[p]
	\centering
	\subfigure[$a_x+a_s=20$]{\includegraphics[width=6.5cm]{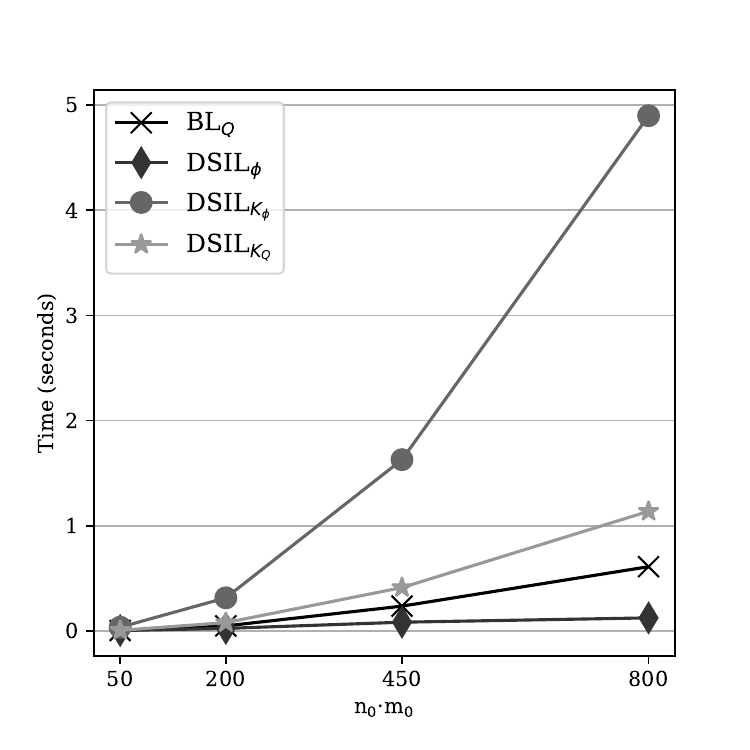}\label{fig:attributesa}}
	\subfigure[$a_x+a_s=200$]{\includegraphics[width=6.5cm]{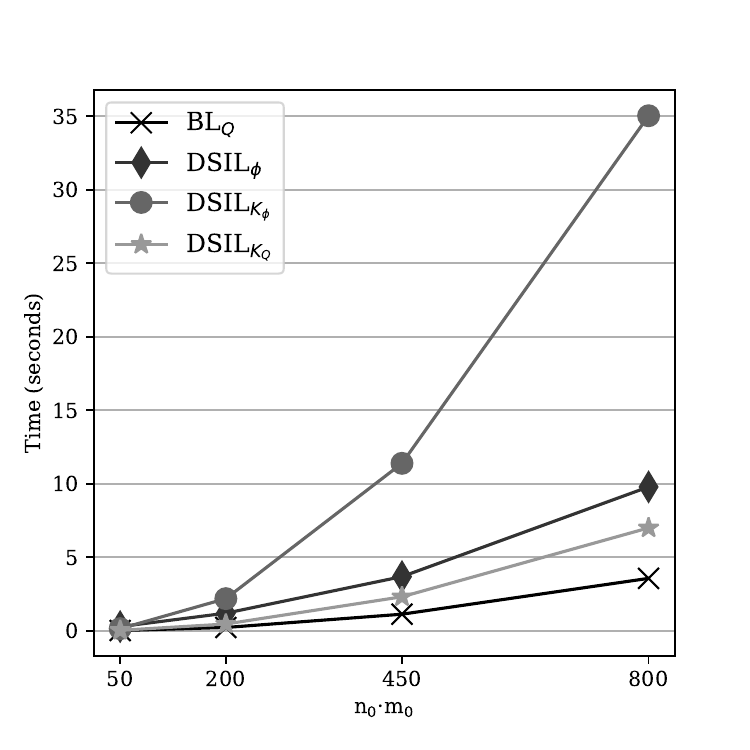}\label{fig:attributesb}}
	\subfigure[$a_x+a_s=500$]{\includegraphics[width=6.5cm]{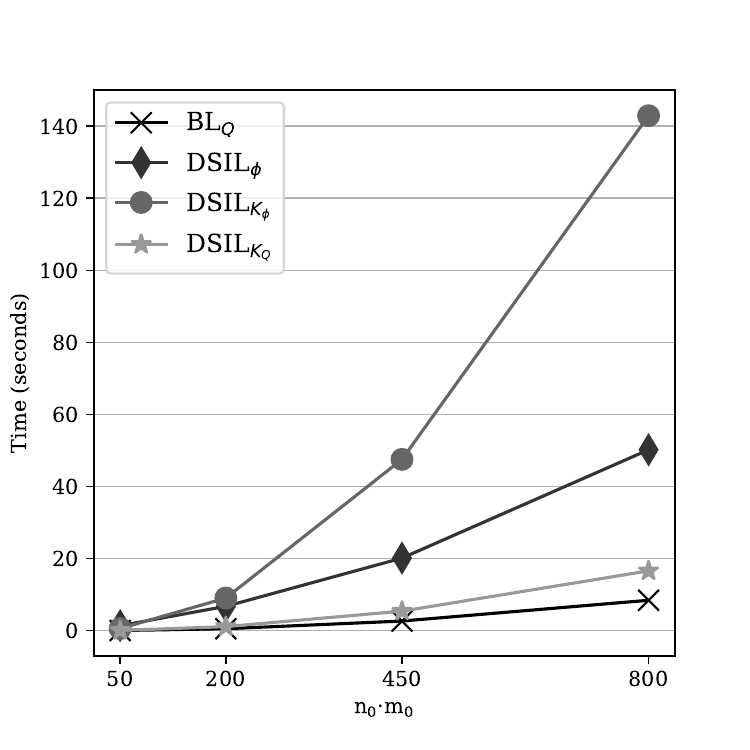}\label{fig:attributesc}}
	\subfigure[$a_x+a_s=1000$]{\includegraphics[width=6.5cm]{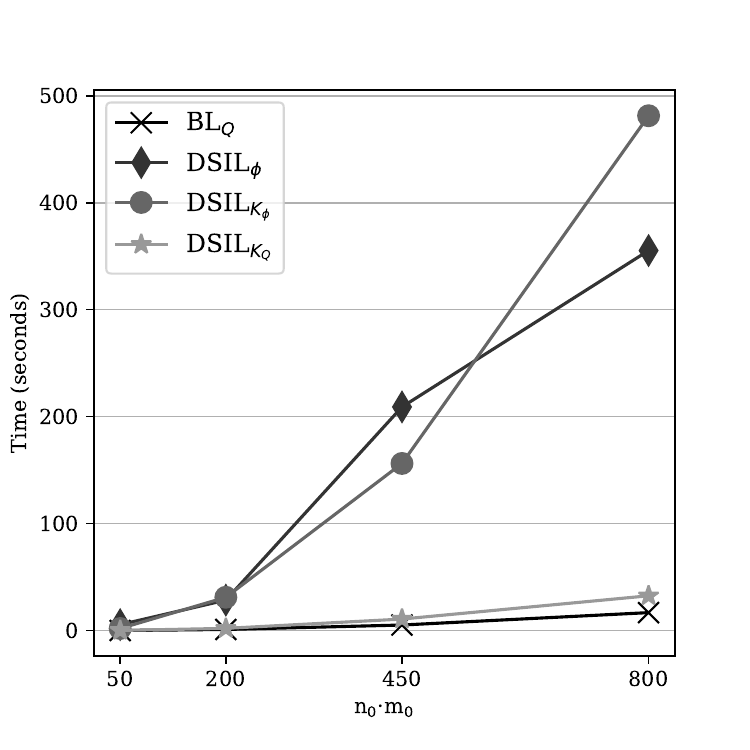}\label{fig:attributesd}}
	\caption{Time (in seconds) of \basep, \dsilm, \dsilk\ and \dsilkq\ when the number of instances that feed the approaches varies and for different values of $a_x+a_s$} \label{fig:attributes}
\end{figure}

\begin{figure}[p]
	\centering
	\subfigure[$n_o\cdot m_o=50$]{\includegraphics[width=6.5cm]{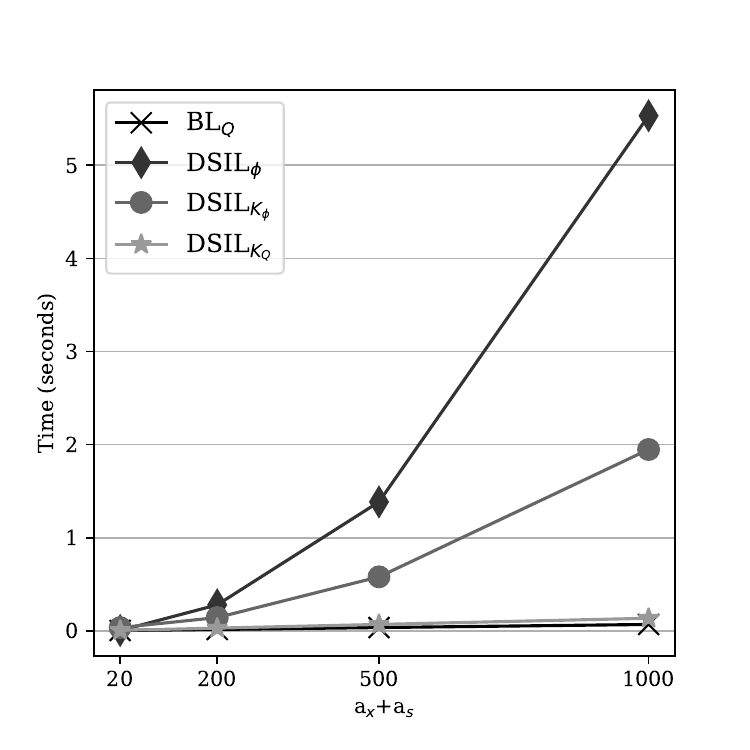}\label{fig:rowsa}}
	\subfigure[$n_o\cdot m_o=200$]{\includegraphics[width=6.5cm]{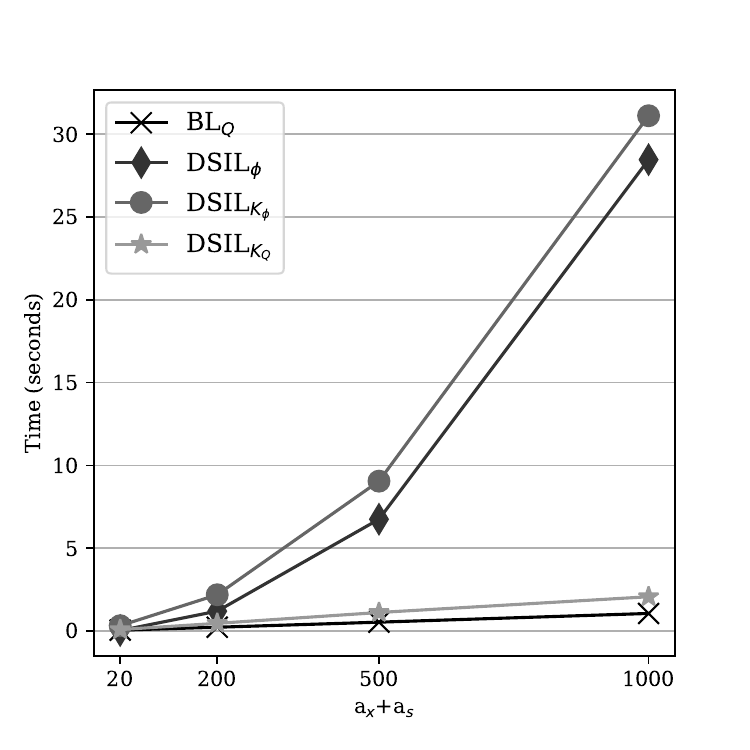}\label{fig:rowsb}}
	\subfigure[$n_o\cdot m_o=450$]{\includegraphics[width=6.5cm]{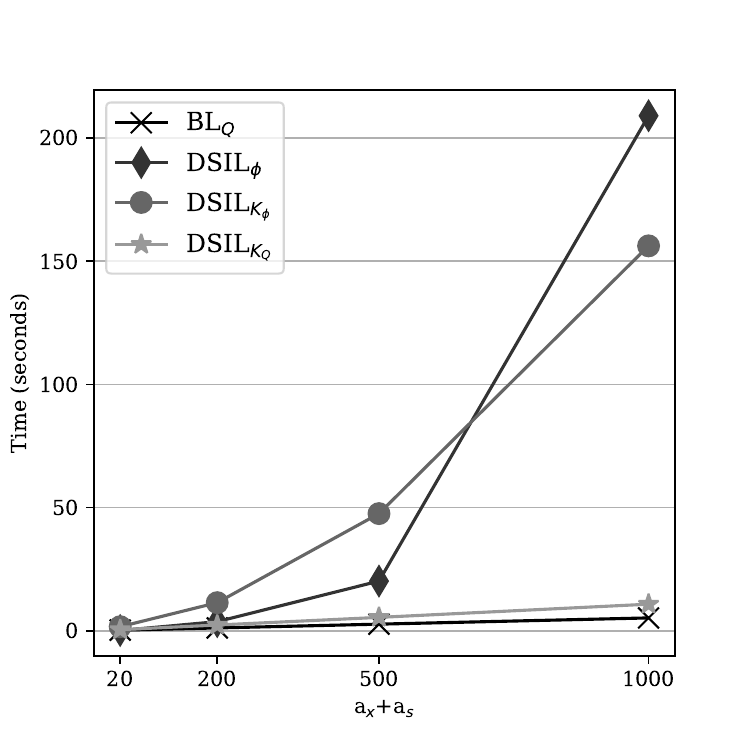}\label{fig:rowsc}}
	\subfigure[$n_o\cdot m_o=800$]{\includegraphics[width=6.5cm]{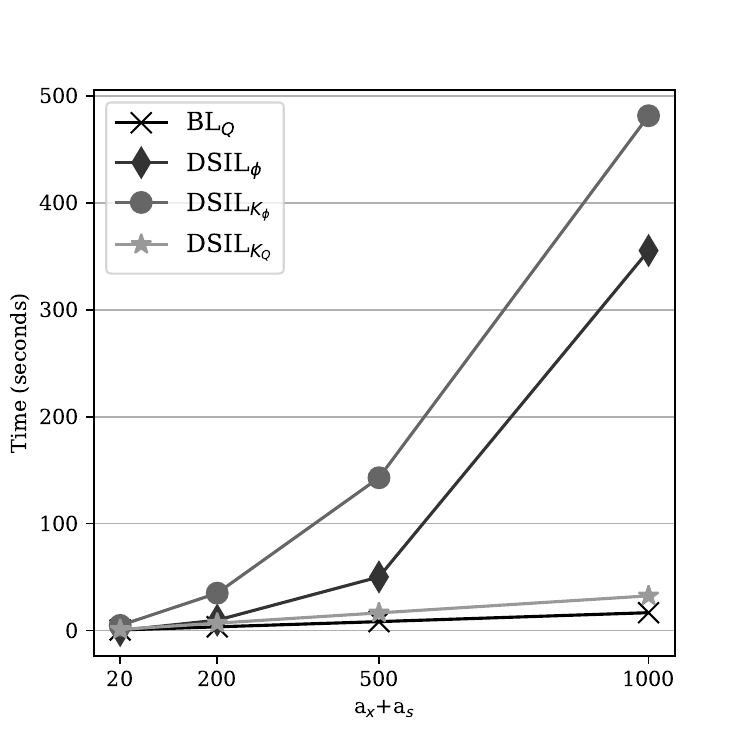}\label{fig:rowsd}}
	\caption{Time (in seconds) of \basep, \dsilm, \dsilk\ and \dsilkq\ when the number of features that feed the approaches varies and for different values of $n_o\cdot m_o$} \label{fig:rows}
\end{figure}

\subsubsection{Computational time analysis of \dsilm, \dsilk\ and \dsilkq}
Figures \ref{fig:attributes} and \ref{fig:rows} show the computational time (in seconds) employed by the three implementations of \dsil\ (\dsilm, \dsilk\ and \dsilkq) over the specific datasets built in order to carry out this analysis. It also shows the computational time of \basep\ as a reference. Particularly, Figure \ref{fig:attributes} shows the computational time varying the number of instances ($n_o\cdot m_o$) for the different values of the number of features ($a_x+a_s$). Conversely, Figure \ref{fig:rows} displays the computational time varying the number of features ($a_x+a_s$) for the different values of the number of instances ($n_o\cdot m_o$). 

On the one hand, \dsilk\ clearly reaches the worst computational time as the number of instances increases (see Figure \ref{fig:attributes}). In fact, this behavior was expected because, as commented before, the interest in using kernels is precisely to avoid computing the kernel via $\phi$ mapping.  On the other hand, dispensing with kernels (\dsilm) is the best option for low values of features, but soon, the computational time starts to rocket as the number of features increases (see the evolution of the computational time of \dsilm\ from Figure \ref{fig:attributesa} to Figure \ref{fig:attributesd}). Indeed, the graph of \dsilm\ is quite closer to \dsilk\ for the highest value of the number of features ($a_x+a_s=1000$). Finally, \dsilkq\ is the steadier option whatever the number of features is. 

Varying the number of features (see Figure \ref{fig:rows}), the implementations of \dsil\ that apply the $\phi$ mapping (\dsilm\ and \dsilk) have a quadratic order computational time, whereas the computational time is of linear order in case of the implementation defined though a linear combination of quadratic kernels (\dsilkq).  Also, \dsilk\ is highly affected by whatever value of the number of instances (see the evolution of the computational time of \dsilk\ from Figure \ref{fig:rowsa} to Figure \ref{fig:rowsd} and especially from $n_o\cdot m_o=50$ to $n_o\cdot m_o=200$).

\section{Conclusions and future work} \label{sec:conclusions}
This paper proposes a one-phase method for zero-shot regression, a task whose goal is to induce models for predicting continuous values for unobserved targets, a kind of targets for which instances have not been collected. Under this lack of instances, zero-shot regression takes advantage of the existence of side information, which is neither features nor targets, but provides information about the relationship between features and targets. The straightforward way of exploiting such information consists of taking them as common features in a unique process. However, some recent approaches have empirically shown that side information deserves to be mined in a different way than the way common instance features are treated. These approaches firstly learn models for the observed targets just taking into account the instance description features and they secondly integrate side information with these models, leading to a two-phase process. The main disadvantage of these approaches is precisely the existence of two different phases, which leads to locally optimize the exploitation of the information instead of as a whole. This paper proposes an alternative in between, that is, it deals with zero-shot regression in a one-phase procedure integrating instance description features and side information simultaneously in order to get a global optimal process, but treating side information properly according to its nature and in any case in a different way that instance features are treated.  The method is firstly defined in terms of the existing but unknown relationship between features and targets and between this relationship and side information. Then, a mapping function is deduced from this definition. However, the high storage requirements of this definition make necessary look for an alternative. For this purpose, the method is then built on the basis of a kernel designed from that mapping function. The main drawback of this kernel is its own definition in terms of the mapping function, which incurs in a high computational cost. In fact, the aim of the kernels is precisely to avoid explicitly applying the mapping function and to perform the inner product in the image space of the mapping function instead. Hence, another alternative definition is provided for the kernel in terms of the existing quadratic kernel, providing an implementation of linear instead of quadratic order with regard to the number of features. 

Several experiments over both artificial and real datasets exhibit the superiority of the novel approach with regard to other recent existing approaches, being statistically significant with regard to some of them. Additional experiments were accomplished to compare the computational time of the different implementations of the new method (the one consisting of directly mapping the instances using and assuming a linear relationship in the image space of the mapping afterwards, the one consisting of defining a kernel via the mapping function and the one consisting of defining the kernel via the existing quadratic kernel). The conclusion is that the computational time is much steadier if the kernel is defined in terms of the existing quadratic kernel both varying the number of instances or features.

In future work, the plan is to extend the kernel design beyond the linear scenario. This means to cope with non-linearity both in the relation between targets and features and between the side information and the observed target model parameters. Another task for future research will be to contemplate a zero-shot multi-regression scenario, where more than one continuous value is predicted simultaneously. 

\section*{Acknowledgments}\label{sec:Acknowledgments}
This research has been partially supported by the Spanish Ministry of Science and Innovation through the grant PID2019-110742RB-I00.

\bibliographystyle{apalike}
\biboptions{numbers}
\bibliography{mybib}

\end{document}